\documentclass[10pt,twocolumn,letterpaper]{article}

\usepackage{iccv}
\usepackage{times}
\usepackage{epsfig}
\usepackage{graphicx}
\usepackage{amsmath}
\usepackage{amssymb}
\usepackage{booktabs}
\usepackage{multirow}
\usepackage{pifont}
\usepackage{xcolor,colortbl}
\usepackage{float}
\usepackage{caption}
\usepackage{subcaption}
\usepackage{easyReview}
\usepackage[accsupp]{axessibility}

\usepackage{algorithm}
\usepackage{algpseudocode}
\usepackage{rotating}

\usepackage[pagebackref=true,breaklinks=true,letterpaper=true,colorlinks,bookmarks=false]{hyperref}

\iccvfinalcopy 

\def\iccvPaperID{2537} 

\ificcvfinal\fi

\begin{document}

\title{Cross-Modal Learning with 3D Deformable Attention for Action Recognition}

\author{Sangwon Kim \quad Dasom Ahn \quad Byoung Chul Ko\thanks{Corresponding author}\\
Keimyung University\\
{\tt\small \{eddiesangwonkim, tommydasomahn\}@gmail.com, niceko@kmu.ac.kr}
}

\maketitle
\ificcvfinal\thispagestyle{empty}\fi

\begin{abstract}
	An important challenge in vision-based action recognition is the embedding of spatiotemporal features with two or more heterogeneous modalities into a single feature. In this study, we propose a new 3D deformable transformer for action recognition with adaptive spatiotemporal receptive fields and a cross-modal learning scheme. The 3D deformable transformer consists of three attention modules: 3D deformability, local joint stride, and temporal stride attention. The two cross-modal tokens are input into the 3D deformable attention module to create a cross-attention token with a reflected spatiotemporal correlation. Local joint stride attention is applied to spatially combine attention and pose tokens. Temporal stride attention temporally reduces the number of input tokens in the attention module and supports temporal expression learning without the simultaneous use of all tokens. The deformable transformer iterates $L$-times and combines the last cross-modal token for classification. The proposed 3D deformable transformer was tested on the NTU60, NTU120, FineGYM, and PennAction datasets, and showed results better than or similar to pre-trained state-of-the-art methods even without a pre-training process. In addition, by visualizing important joints and correlations during action recognition through spatial joint and temporal stride attention, the possibility of achieving an explainable potential for action recognition is presented.
\end{abstract}

\section{Introduction}
\label{sec:intro}

\begin{figure}
	\centering
	\includegraphics[width=0.9\linewidth]{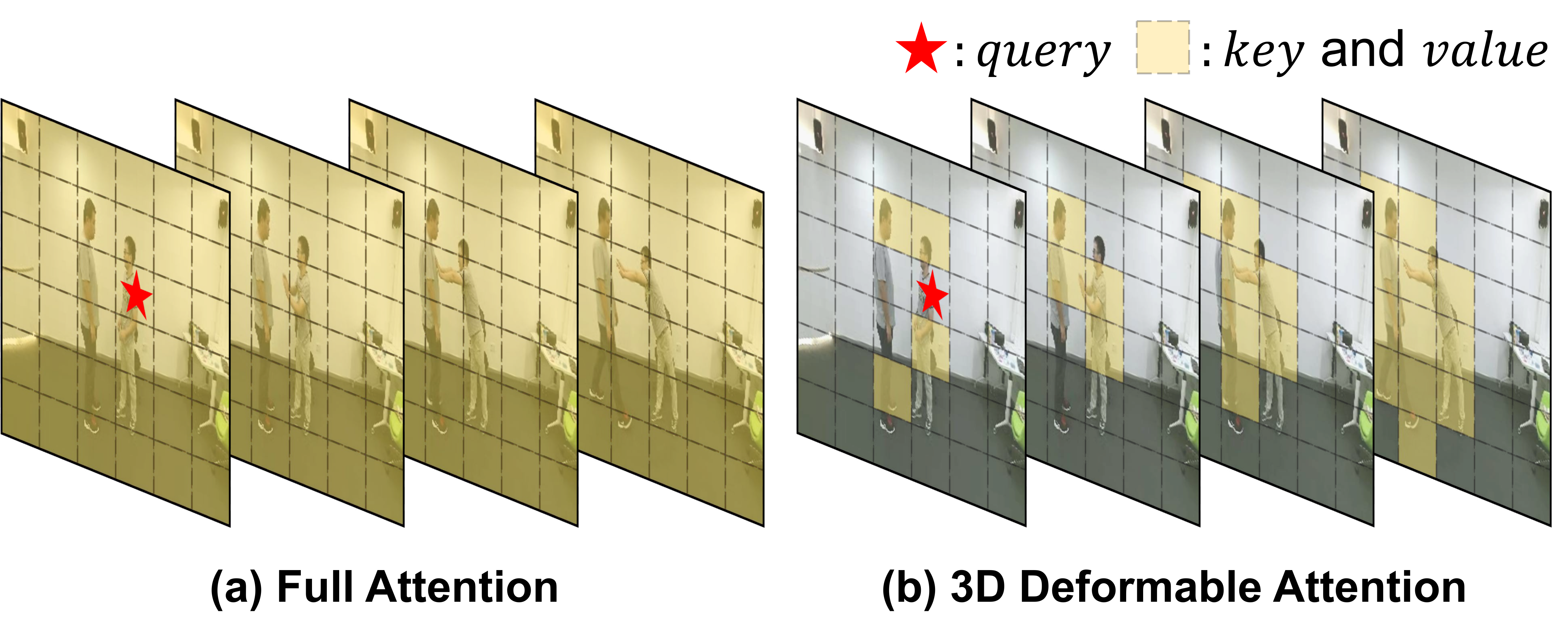}
	\caption{\textbf{Comparison between (a) Full attention and (b) the proposed 3D deformable attention.} Full attention considers all tokens against a specific \textit{query} in a complete sequence. By contrast, 3D deformable attention considers only intense tokens with adaptive receptive fields.}
	\label{fig1}
\end{figure}

Spatiotemporal feature learning is a crucial part of action recognition, which aims to fuse not only the spatial features of each frame but also the temporal correlation between input sequences. Previous studies on action recognition \cite{posec3d, mmnet, tsmf, c3d, i3d, s3d} investigated the application of 3D convolutional kernels with an additional temporal space beyond the 2D spatial feature space. Since then, 3D convolutional neural networks (CNN) have achieved a promising performance and have eventually become the \textit{de facto} standard for various action recognition tasks using sequential data. Vision transformers (ViTs) for action recognition, which have peaked in popularity, have recently been used to explore a 3D token embedding to fuse the temporal space within a single token. However, ViTs-based action recognition methods \cite{star, trans4soar} are limited in that they can only conduct spatiotemporal feature learning within restricted receptive fields.

To avoid this problem, several studies \cite{dcn, defdetr,dat} have been conducted to allow more flexible receptive fields for deep learning models. Deformable CNN leverages dynamic kernels to capture the intense object regions. First, they determine the deformable coordinates using embedded features. The kernel is then applied to the features extracted from the deformable coordinates. Deformable ViTs \cite{dat, defdetr} encourage the use of an existing attention module to learn deformable features. The \textit{query} tokens are projected onto the coordinates to obtain deformable regions from the \textit{key} and \textit{value} tokens. The deformed value tokens are then applied to the attention map, which is generated through a scaled dot product of the input \textit{query} and deformed \textit{key} tokens. These methods suggest a new approach that can overcome the limitations of existing standardized feature learning. However, despite some impressive results, these studies are still limited in that they are only compatible with the spatial dimensions. Therefore, as a primary challenge, there is a need for the development of novel and deformable ViTs that can learn spatiotemporal features from image sequences.

Another challenge is the efficient application of multimodal input features to an action recognition model. Action recognition is classified into three categories based on the feature type. The first is a video-based approach \cite{actmachine, glimpse, actnet, nflb, x3d, cscnn, vtn}, which has traditionally been used for action recognition. This approach is limited by a degraded performance caused by noise, such as varying object sizes, occlusions, or different camera angles. The second is a skeleton-based approach \cite{stgcn, asgcn, shiftgcn, infogcn, ctrgcn}, which mainly converts poses into graphs for recognizing actions through a graph neural network (GNN). Although this approach is robust against noise, its performance is highly dependent on the pose extraction method. To overcome the shortcomings of the previous two approaches, the third method aims to simultaneously fuse heterogeneous domain features using multimodal or cross-modal learning. With this approach, video and skeleton features are jointly trained simultaneously. However, because most related studies use a separate model composed of a GNN + CNN or CNN + CNN for each modality, there is a limit in constructing an effective single model.

To alleviate the drawbacks stated above, we propose the use of transformer with 3D deformable attention for dynamically utilizing the spatiotemporal features for action recognition. In this way, the proposed model applies flexible cross-modal learning, which handles the skeletons and video frames in a single transformer model. The skeletons are projected onto sequential joint tokens, and each joint token contains an activation at every joint coordinate. To provide effective cross-modal learning between each modality, the proposed method adopts a cross-modal token that takes the role of mutually exchanging contextual information. Therefore, the proposed model is capable of achieving a boosted performance without an auxiliary submodel for the cross-modalities. Figure \ref{fig1} shows a comparison between the previous full attention and the proposed 3D deformable attention. In the case of the full attention shown in Fig. \ref{fig1} (a), all tokens within a spatiotemporal space are covered against a specific \textit{query} token. By contrast, our proposed 3D deformable attention scheme, shown in Fig. \ref{fig1} (b), considers only tokens with high relevance within the entire spatiotemporal space. The main contributions of this study are as follows:

\begin{itemize}
	\item We propose the first 3D deformable attention that adaptively considers the spatiotemporal correlation within a transformer as shown in Fig. \ref{fig1} (b), breaking away from previous studies that consider all tokens against a specific \textit{query} in a complete sequence.
	\item We propose a cross-modal learning scheme based on complementary cross-modal tokens. Each cross-modal token delivers contextual information between the different modalities. This approach can support a simple yet effective cross-modal learning within a single-transformer model structure.
	\item We present qualitative evidence for 3D deformable attention with visual explanations and prove that the proposed model outperforms several previous state-of-the-art (SoTA) methods.
\end{itemize}

\section{Related Works}
\label{sec:relay}

\begin{figure*}[htp]
	\centering
	\includegraphics[width=0.9\linewidth]{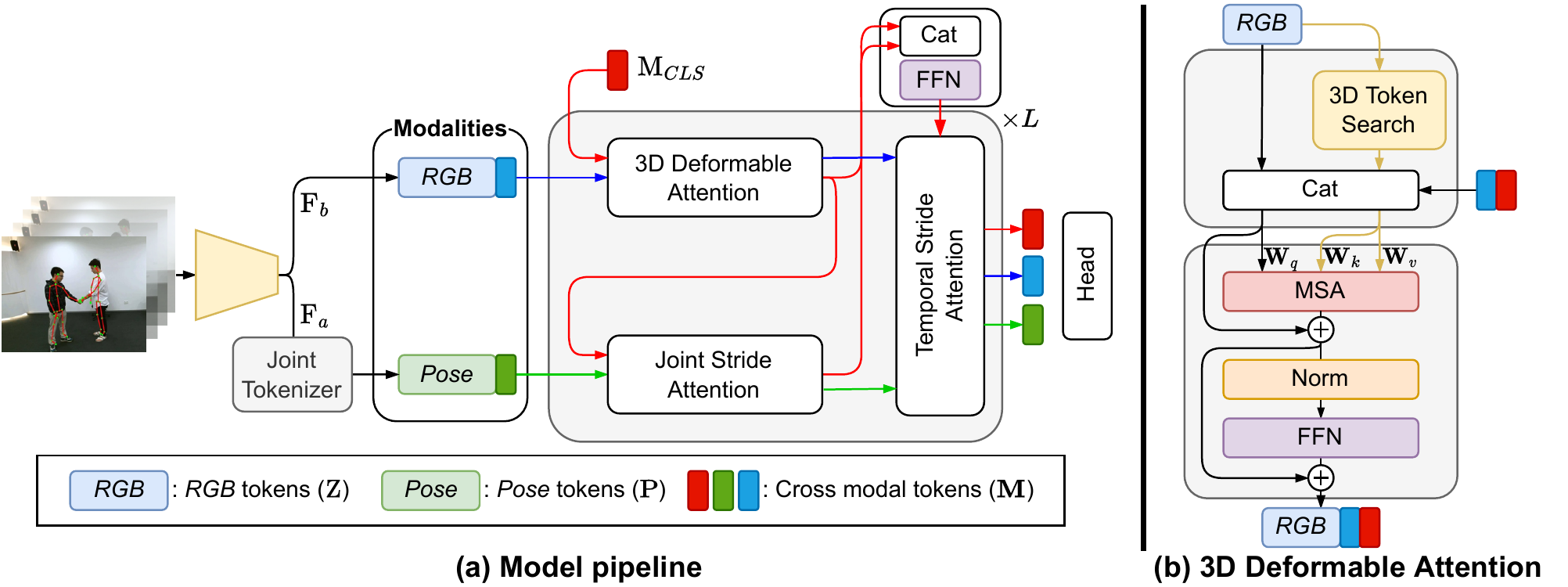}

	\caption{\textbf{Overview of our 3D deformable transformer.} (a) Our model consists of a backbone and a series of transformer blocks. Each transformer block uses different modality tokens to process intrinsic properties from various domains and fuse the modalities with cross-modal tokens. The proposed model includes joint stride and temporal stride attention to reduce computational costs. (b) The proposed 3D deformable attention includes the 3D token search (3DTS) and the attention block. The input \textit{RGB} token $\mathrm{Z}$ is embedded as $query$ token after concatenating with modal tokens. The deformable token from the 3DTS is also concatenated with modal tokens, then it is multiplied with \textit{key} ($\mathbf{W}_k$) and \textit{value} ($\textbf{W}_v$) weights. These are then fed to the multi-head self-attention (MSA) to interact with the \textit{query} token.}
	\label{fig2}
\end{figure*}

\noindent
\textbf{Spatiotemporal learning for action recognition.} Early studies in this area focused primarily on employing a 3D CNN, which is an extension of a 2D CNN. This has become a central remedy in vision-based action recognition in recent years. PoseC3D \cite{posec3d} combines 3D volumetric heatmaps from the skeletons and frames of the input video. SlowFast \cite{slowfast} makes a significant contribution to the field by providing a frame-fusion scheme between different frame rates. There are also related methods \cite{3dcnnaction, c3d,cscnn, closer,i3d, s3d,  glimpse, x3d,  stda} that explore the use of a 3D CNN architecture for action recognition. STDA \cite{stda} applies a 3D deformable CNN that captures substantial intense regions for spatiotemporal learning. Over the last few years, focus has shifted toward skeleton-based action recognition with respect to the emergence of a GNN. ST-GCN \cite{stgcn} has become a baseline adopting separate spatial and temporal representation modules for spatiotemporal modeling. In addition, ViTs have attracted considerable attention owing to their superior performance in sequential tasks. STAR \cite{star} applies cross-attention for the fusing of temporal correlations between spatial representations. ViViT \cite{vivit} embeds an input video with a 3D tokenizer to compose the spatiotemporal features in a single token. Other studies \cite{stmixing, star, videoswintr} have adopted a temporal stride to capture the diversity between different time steps. However, the concept of a 3D deformation, despite its excellent performance, cannot be applied to the attention of ViTs owing to various structural constraints.

\noindent
\textbf{Cross-modal learning for action recognition.} Most current action recognition methods use various modalities with video frames and skeletons. Several methods \cite{vpn, mmnet, tsmf, vpnpp} employ a graph convolutional network (GCN) to handle a raw skeleton input and a CNN for the video frames. VPN \cite{vpn} applies GCN subnetworks to support the CNN. The footages of GCN networks are linearly combined with the CNN feature maps. MMNet \cite{mmnet} introduced a multimodal network with two GCN subnetworks and CNNs. Each subnetwork embeds the features separately, and these features are then summed at the end of the network. Other studies \cite{potion, pa3d, dynamicmotion, star, posec3d, harheatmap} transformed graphical skeletons into the heatmaps. PoseC3D \cite{posec3d} uses dual 3D CNN branches for video frames and 3D volumetric heatmaps. It does not explicitly consider the spatial relationships between joints in the skeleton. This may limit its ability to capture complex and subtle movements, or to distinguish between similar actions that involve different joint configurations. STAR \cite{star} proposed joint tokens generated by combining CNN feature maps with 2D joint heatmaps. To fuse the two modalities, they concatenated multiclass tokens by combining different modal tokens. Despite the improved performance of cross-modal learning, video frames and skeleton modalities are merely integrated, thus neglecting a careful design. We propose an effective feature fusion method called a cross-modal token. To exchange contextual information, each token is dispatched to another modality.

\noindent
\textbf{Transformer with deformable attention.} The idea of a 2D deformable CNN for the learning of deformable features has been applied to the attention module of a ViT, achieving an excellent performance in various applications, including image classification. Deformable DETR \cite{defdetr} was applied to object detection and demonstrated its ability to accurately detect objects of various sizes. A deformable attention transformer (DAT) \cite{dat} with an improved numerical stability and a robust performance was recently proposed. In terms of action recognition, 3D deformable CNNs \cite{stda, deformabletube} for spatiotemporal learning showed a better performance than a 2D deformable CNN but were not applied to a transformer owing to the structural constraints of an attention optimized for spatial feature embedding.
Therefore, in this study, we propose a new 3D deformable transformer capable of fusing cross-modal features using cross-modal tokens. The proposed deformable method enables 3D deformable feature embedding based on local joint stride and temporal stride attention.
The remainder of this paper is organized as follows. Section \ref{sec:approach} provides a detailed explanation of the proposed approach. Section \ref{sec:exp} provides an experimental analysis of several benchmarks as well as visual descriptions. Finally, Section \ref{sec:conclusion} provides some concluding remarks regarding this research.

\section{Approach}
\label{sec:approach}

We propose a 3D deformable transformer for action recognition with adaptive spatiotemporal receptive fields and a cross-modal learning scheme. The overall architecture of the proposed model is shown in Fig. \ref{fig2} and is described in detail in the following sections.

\begin{figure}[t!]
	\centering
	\includegraphics[width=0.8\linewidth]{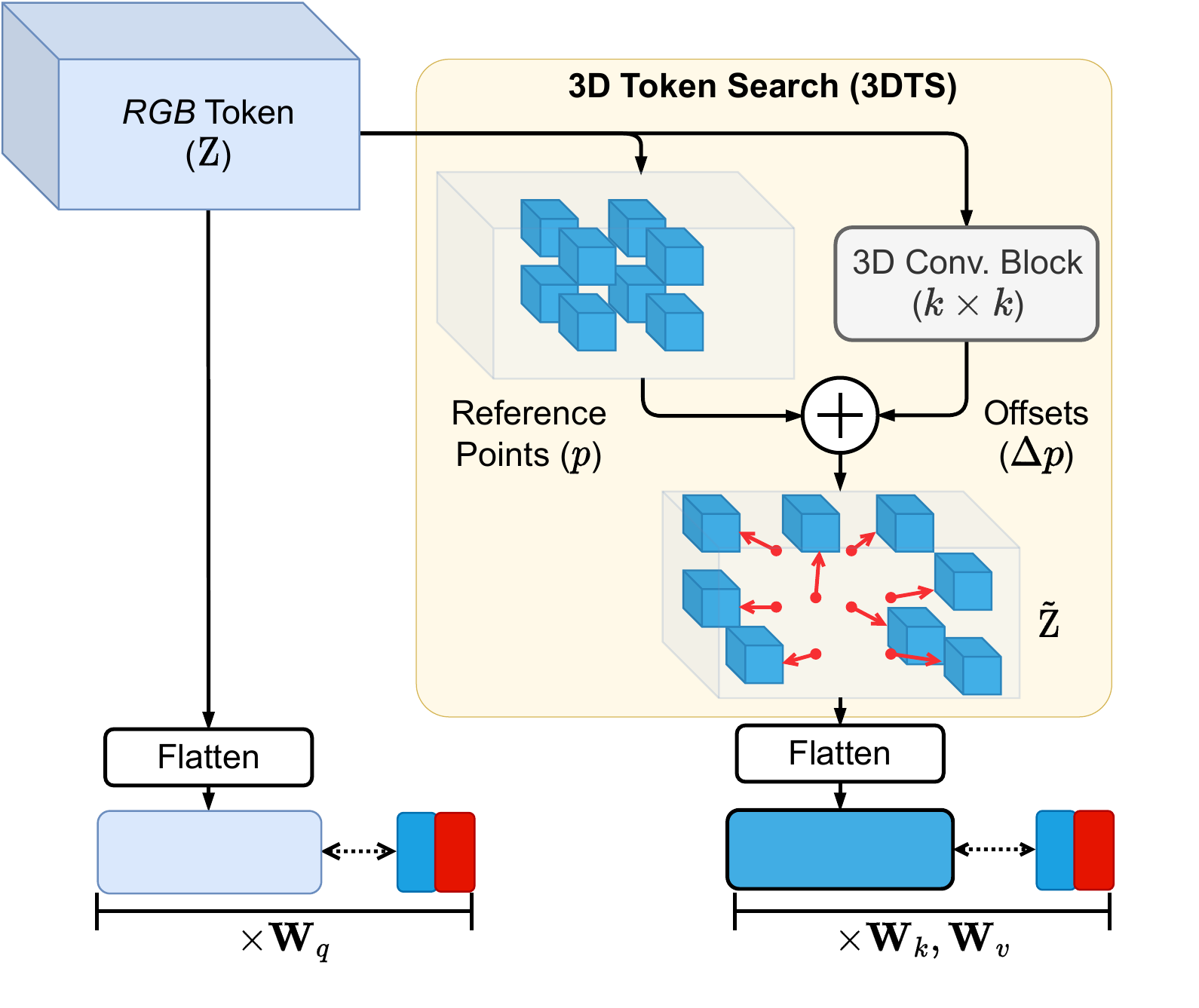}

	\caption{\textbf{Illustration of proposed 3D deformable attention for adaptive spatiotemporal learning.} Using 3DTS, an offset vector ($\Delta p$) from a 3D conv block finds the deformed tokens by moving the reference points ($p$) scattered on the input \textit{RGB} tokens $\mathrm{Z}$. The 3D deformable tokens $\tilde{\mathrm{Z}}$ have the same number of tokens as the reference points ($p$).}
	\label{fig3}
\end{figure}

\subsection{Cross-modal learning}

In action recognition, cross-modal learning has been the mainstream, leveraging various modalities such as video frames and skeletons. Several successful studies \cite{posec3d, mmnet, tsmf, vpn, mmtm, vpnpp} have employed subnetworks that handle different domain features. However, these designs eventually increase the redundancy and complexity owing to the domain-specific subnetworks. We propose simple yet effective cross-modal learning for mutually exchanging contextual information. Our cross-modal learning method consists of a backbone \cite{mcresnet}, which provides intermediate feature maps and sequential tasks. When the image has a height  $H$, width $W$, temporal dimension $T$, and feature dimension $C$, the backbone network feeds visual feature maps $\mathrm{F}_a \in \mathbb{R}^{C \times T \times \frac{H}{2} \times \frac{W}{2}}$ and $\mathrm{F}_b \in \mathbb{R}^{4C \times T \times \frac{H}{8} \times \frac{W}{8}}$, extracted from intermediate layers. In the case of $\mathrm{F}_b$, we consider it as the \textit{RGB} modality input for visual representation learning, whereas the local level feature map, $\mathrm{F}_a$, is regarded as the \textit{pose} modality input by combining with skeletons. To fuse both modalities, we apply the following concepts:

\noindent
\textbf{\textit{Pose} modality.} To design a cross-modal learning scheme with an alleviated redundancy, we propose visual feature-oriented \textit{pose} tokens combined with joint heatmaps, such as \cite{posec3d, star}. First, the sequential skeletons are decomposed into single-joint units. Each joint is then recomposed into a joint heatmap $\mathcal{H} \in \mathbb{R}^{T \times R \times \frac{H}{2} \times \frac{W}{2}}$ by projecting the joints toward an empty voxel at the corresponding coordinates $(x_{t,r},y_{t,r})$. Here, $R$ is the number of joints, and spatial dimension follows feature map size of $\mathrm{F}_a$. Finally, \textit{pose} tokens $\mathrm{P}$ using the joint tokenizer shown in Fig. \ref{fig2} (a) using the multiplication of $\mathrm{F}_a$ and the Gaussian blur output at time $t$ as follows:

\begin{equation}
	\mathrm{P}_t = ||_r \sum^{\frac{H}{2}}_{j} \sum^{\frac{W}{2}}_{i} \mathrm{F}_{a,t}(i,j)\mathcal{H}_{t,r}(i,j)
	\label{eq1}
\end{equation}

\begin{equation}
	\mathcal{H}_{t,r}(i,j) = e^{-\frac{(i-x_{t,r})^2+(j-y_{t,r})^2}{2\sigma^2}}
	\label{eq2}
\end{equation}

\noindent
where $\mathrm{P} \in \mathbb{R}^{C \times T \times R}$ consists of $R$ \textit{pose} tokens for every skeleton sequence with $C$ feature dimensions. $||$ indicates concatenation. To meet the feature dimensions with \textit{RGB} modality, $\mathrm{F}_b$, linear projection is applied to \textit{pose} tokens, resulting in $\mathrm{P} \in \mathbb{R}^{4C \times T \times R}$.

\noindent
\textbf{\textit{RGB} modality.} With \textit{RGB} modality, the extracted visual feature map $\mathrm{F}_b$ is regarded as \textit{RGB} tokens $\mathrm{Z} \in \mathbb{R}^{4C \times T \times \frac{H}{8} \times \frac{W}{8}}$ and is fused with the \textit{position embedding}.

\subsection{3D deformable transformer}

\noindent
\textbf{Cross-modal tokens.} An intuitive method is to concatenate all tokens from both modalities, considering the characteristics of each token, and then combining information through the transformer stacks. However, to combine different modalities in a single transformer, a deliberate design is required, and the modalities must be cooperative and complementary. Similarly, in STAR \cite{star}, the authors employ multi-class tokens for cross-modal learning. Despite a simple yet effective approach, on par with other transformers, it is aimed only at an information fusion for all tokens without considering the intrinsic properties and complementarities of various modalities. Therefore, we propose a cross-modal token that effectively combines the different modalities within the transformer. The cross-modal token $\mathbf{M} \in \mathbb{R}^{4C \times T \times 3}$ is a set of three trainable tokens: \textit{CLS}, \textit{RGB} and \textit{pose} modal tokens. In previous studies \cite{vit, deit}, \textit{CLS} tokens were used as the final embeddings  fusing information by interacting with other tokens. We consider the \textit{CLS} token $\mathrm{M}_{CLS} \in \mathbb{R}^{4C \times T \times 1}$ as a `modality-mixer' compiling the remaining two modal tokens, which are dispatched to mutual modalities to trade their domain knowledge. The first $\mathrm{M}_{RGB}$ and $\mathrm{M}_{CLS}$ tokens are fed to the 3D deformable attention. Then, the output \textit{RGB} and \textit{CLS} modal tokens, $\mathrm{M}_{RGB}$ and $\mathrm{M}_{CLS}$ of a 3D deformable attention, reflect information from their own domains through separated transformer blocks cooperating with the dispatched \textit{CLS} tokens. Hereafter, we introduce the 3D deformable attention shown in Fig. \ref{fig2} (b), which is the core of the proposed transformer.

\noindent
\textbf{3D deformable attention.} Although transformers have recently become a new standard in vision tasks, relatively few studies have been conducted on action recognition tasks. Because the nature of a transformer considers long-term relations between the input tokens, it may lead to an exponentially increasing computational complexity with the time steps. In addition, to solve the problem of static transformers, DAT \cite{dat} that flexibly selects the \textit{key} and \textit{value} positions in a self-attention has been proposed; however, it is unsuitable for an action recognition that has to deal with cross-modalities and spatiotemporal features. To alleviate the complexity while maintaining the nature of the transformer, inspired by \cite{dat}, we propose the use of 3D deformable attention for action recognition, as shown in Fig. \ref{fig2} (b). 3D deformable attention can adaptively capture spatiotemporal features on the \textit{RGB} modality.

The 3D deformable attention module consists of a 3D token search (3DTS) and multi-head self-attention (MSA) with a feed-forward network (FFN), as shown in Fig. \ref{fig2} (b). First, the input of the module, \textit{RGB} token  $\mathrm{Z}$, is fed to the 3DTS, which contains a two-layered Conv3D with kernel $k$. After the first Conv3D, layer normalization (LN) and GELU non-linearity are applied. The last Conv3D generates offsets ($\Delta p$) that contain flow fields against the reference points ($p$). The reference points are defined as being regularly scattered within a 3D space. The offsets guide the reference points to find discriminative token coordinates in the spatiotemporal tokens  $\mathrm{Z}$, as shown in Fig. \ref{fig3}. 3D deformable tokens $\tilde{\mathrm{Z}}$ are configured by selecting the tokens from the adjusted coordinates taken from the offsets.

\begin{equation}
	\tilde{\mathrm{Z}} = \mathrm{3DTS}(\mathrm{Z} ; \omega)
	\label{eq3}
\end{equation}

\noindent
where $\mathrm{Z} \in \mathbb{R}^{4C \times T \times \frac{H}{8} \times \frac{W}{8}}$ and $\tilde{\mathrm{Z}} \in \mathbb{R}^{4C \times \tilde{T} \times \tilde{H} \times \tilde{W}}$ are the input and selected \textit{RGB} tokens respectively. The size of $\tilde{T}$, $\tilde{H}$ and $\tilde{W}$ are determined based on the kernel size $k$. In our case, we set the $k$ as 7 without padding for sparsely extracting deformable tokens and increasing the efficiency. In addition, $\mathbf{W}_q\in\mathbb{R}^{4C \times 4C}$ and $\omega$ are trainable weight and model parameters for the MSA and 3D conv block in the 3DTS, respectively. It should be noted that while \textit{query} tokens are composed in the same manner as the transformer, the \textit{key} and \textit{value} tokens are composed of selected tokens from the 3DTS. Further details on our implementation are in Appendix \ref{app:b}.

These tokens are then embedded into the \textit{key} and \textit{value} tokens using $\mathbf{W}_k$ and $\mathbf{W}_v$, respectively. Herein, we aim to make the  $\mathrm{M}_{RGB}$ token faithfully learn the \textit{RGB} modality features, and $\mathrm{M}_{CLS}$ trades domain knowledge between the \textit{RGB} and \textit{pose} modalities. To fuse cross-modal tokens with the \textit{RGB} modality, three tokens,  $\mathrm{M}_{RGB}$, $\mathrm{M}_{CLS}$ and spatiotemporal feature tokens $\mathrm{Z}$, are concatenated to token $\mathbf{X}$.

\begin{equation}
	\mathbf{X} = [\mathrm{Z}||\mathrm{M}_{RGB}||\mathrm{M}_{CLS}]
	\label{eq4}
\end{equation}

\noindent
where $\mathrm{M}_{RGB}$ and $\mathrm{M}_{CLS}$ are obtained from a portion of the proposed cross-modal tokens representing the \textit{RGB} modality and modality head, respectively.

Similarly, the selected deformable tokens $\tilde{\mathrm{Z}}$ are coupled with two cross-modal tokens to produce $\tilde{\mathbf{X}}$.

\begin{equation}
	\tilde{\mathbf{X}} = [\tilde{\mathrm{Z}}||\mathrm{M}_{RGB}||\mathrm{M}_{CLS}]
	\label{eq5}
\end{equation}

\noindent
Then, $\mathbf{X}$ is multiplied with \textit{query} weight $\mathbf{W}_q$ and, $\tilde{\mathbf{X}}$ is multiplied with \textit{key} and \textit{value} weights, $\mathbf{W}_k$ and $\mathbf{W}_v$, respectively. Those recomposed tokens are fed into multi-head self-attention as a \textit{query}, \textit{key} and \textit{value} for each.

\begin{equation}
	\mathbf{X} = \mathbf{X} + \mathrm{MSA}(\mathbf{X\mathbf{W}}_q, \mathbf{\tilde{X}\mathbf{W}}_k, \mathbf{\tilde{X}\mathbf{W}}_v)
	\label{eq6}
\end{equation}

\noindent
The output $\mathbf{X}$ of 3D deformable attention is finally obtained by applying a LN and FFN.

\begin{equation}
	[\mathrm{Z}, \mathrm{M}_{RGB}, \mathrm{M}_{CLS}^{'}] = \mathbf{X} + \mathrm{FFN}(\mathrm{LN(\mathbf{X})})
	\label{eq7}
\end{equation}

We visualized the attention scores of the selected tokens from the proposed 3D deformable attention shown in Fig. \ref{fig7}. As indicated in Fig. \ref{fig7}, our proposed 3DTS faithfully identifies the fundamental intense regions with adaptive receptive fields against entire sequences.

\begin{figure}
	\centering
	\includegraphics[width=0.9\linewidth]{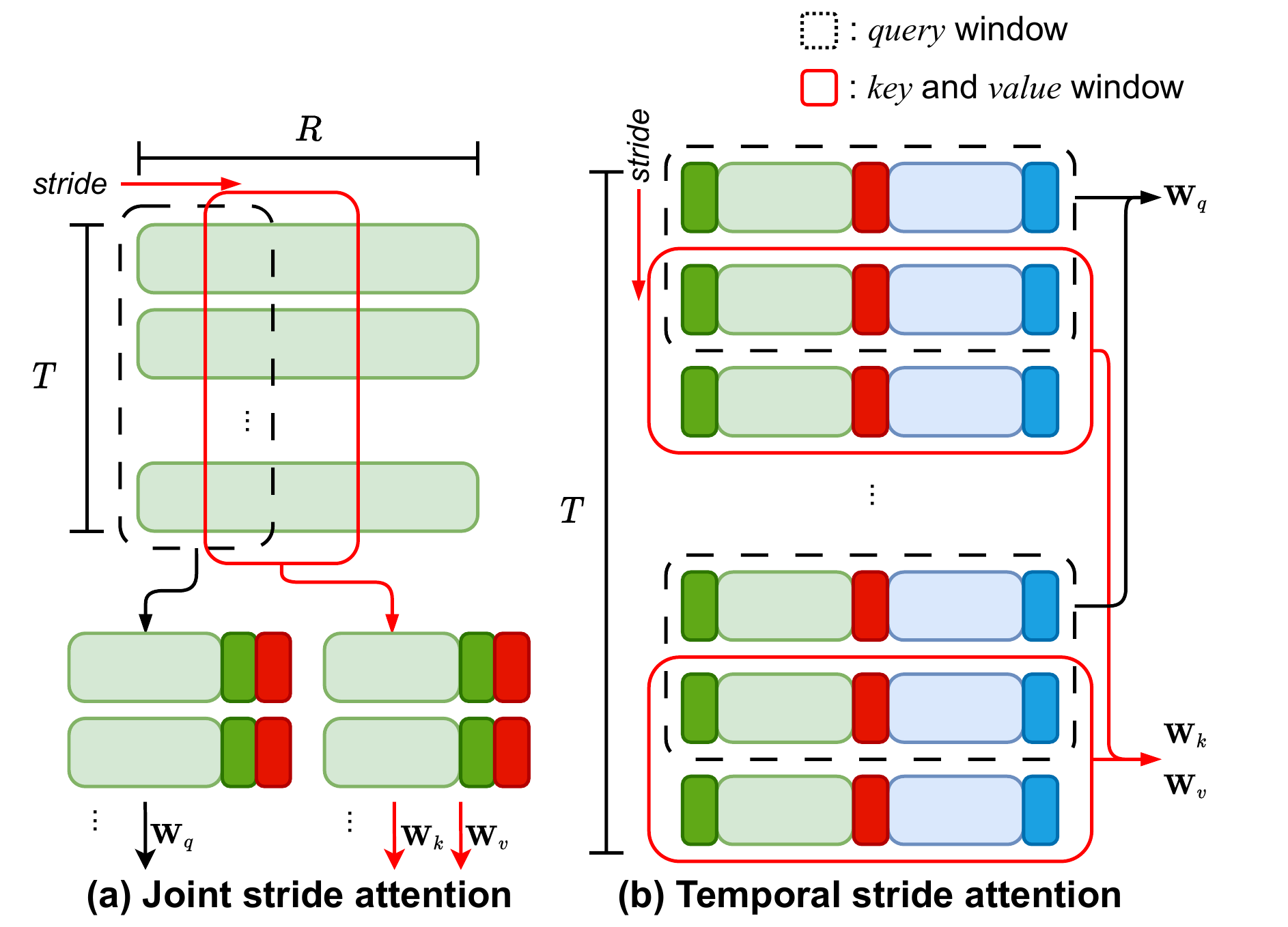}
	\caption{\textbf{Proposed stride attention modules.} (a) Joint stride attention, where a series of joint tokens are grouped into \textit{query}, \textit{key}, and \textit{value} including all temporal dimensions with a stride window. (b) Temporal stride attention where tokens are bundled with temporal stride windows to fuse the changes with the time steps.}
	\label{fig4}
\end{figure}

\noindent
\textbf{Local joint stride attention.} In action recognition, there are often multiple people appearing in a scene; therefore, the number of joint tokens increases with the number of people. To reduce the computational complexity, we concatenate the joints of multiple people into a series of joint tokens. Although this approach is an efficient way to process multiple people in the same scene simultaneously without significantly increasing the complexity, it still results in a problem in that the size of the joint token increases exponentially as the number of people increases. To avoid this problem, we configure the \textit{query}, \textit{key}, and \textit{value} tokens using a sliding window on the joint tokens, as shown in Fig. \ref{fig4} (a). All tokens in each sliding window are flattened and then concatenated with $\mathrm{M}_{pose}$ and $\mathrm{M}_{CLS}^{'}$ dispatched from a 3D deformable attention to apply a scaled-dot product. This is more efficient than calculating all tokens at once and maintaining the relations with each other. The output of the joint stride attention is the \textit{pose} token $\mathrm{P}$ and modal tokens $\mathrm{M}_{pose}$ and $\mathrm{M}_{CLS}^*$.

The calculated \textit{RGB} tokens $\mathrm{Z}$ and \textit{pose} tokens $\mathrm{P}$ are fed to the temporal stride attention module. Before this step, to fuse contextual information from each modality, $\mathrm{M}_{CLS}^{'}$ memorized from the 3D deformable attention and $\mathrm{M}_{CLS}^*$ calculated from the joint stride attention are projected together into a new single $\mathrm{M}_{CLS}$ as shown in Fig. \ref{fig2} (a). Subsequently, the temporal stride attention module learns the correlations between temporal changes against tokens concatenated with cross-modal tokens.

\begin{table*}
	\centering
	\caption{Accuracy comparisons with SoTA approaches on NTU60 and NTU120. P and R denote \textit{pose} and \textit{RGB} modalities, respectively. $\dag$ indicates estimated \textit{pose}.}
	\begin{tabular}{llccccc}
		\toprule
		\multirow{2}{*}{\textbf{Method}}      & \multirow{2}{*}{\textbf{Modality}} & \multirow{2}{*}{\textbf{Pre-trained}} & \multicolumn{2}{c}{\textbf{NTU60}} & \multicolumn{2}{c}{\textbf{NTU120}}                                 \\ \cmidrule{4-5}  \cmidrule{6-7}
		                                      &                                    &                                       & XSub (\%)                          & XView(\%)                           & XSub (\%)     & XSet(\%)      \\
		\midrule
		3s-AimCLR \cite{aimclr}               & P                                  & \ding{51}                             & 86.9                               & 92.8                                & 80.1          & 80.9          \\
		PoseC3D \cite{posec3d}                & P$^\dag$                           & \ding{51}                             & 93.7                               & 96.6                                & 86.0          & 89.6          \\
		VPN \cite{vpn}                        & R+P                                & \ding{51}                             & 95.5                               & 98.0                                & 86.3          & 87.8          \\
		VPN++ (3D \textit{pose}) \cite{vpnpp} & R+P                                & \ding{51}                             & 96.6                               & 99.1                                & 90.7          & 92.5          \\
		MMNet \cite{mmnet}                    & R+P                                & \ding{51}                             & 96.0                               & 98.8                                & 92.9          & 94.4          \\
		PoseC3D \cite{posec3d}                & R+P$^\dag$                         & \ding{51}                             & 97.0                               & 99.6                                & 95.3          & 96.4          \\
		\midrule
		ST-GCN	\cite{stgcn}                   & P                                  &                                       & 81.5                               & 88.3                                & -             & -             \\
		DualHead-Net	\cite{dualheadnet}       & P                                  &                                       & 92.0                               & 96.6                                & 88.8          & 89.3          \\
		Skeletal GNN	\cite{skeletalgnn}       & P                                  &                                       & 91.6                               & 96.7                                & 87.5          & 89.2          \\
		CTR-GCN	\cite{ctrgcn}                 & P                                  &                                       & 92.4                               & 96.8                                & 88.8          & 90.6          \\
		InfoGCN	\cite{infogcn}                & P                                  &                                       & 93.0                               & 97.1                                & 89.8          & 91.2          \\
		KA-AGTN	\cite{kaagtn}                 & P                                  &                                       & 90.4                               & 96.1                                & 86.1          & 88.0          \\
		PoseMap	\cite{posemap}                & R+P$^\dag$                         &                                       & 91.7                               & 95.2                                & -             & -             \\
		MMTM \cite{mmtm}                      & R+P                                &                                       & 91.9                               & -                                   & -             & -             \\
		STAR \cite{star}                      & R+P                                &                                       & 92.0                               & 96.5                                & 90.3          & 92.7          \\
		TSMF \cite{tsmf}                      & R+P                                &                                       & 92.5                               & 97.4                                & 87.0          & 89.1          \\
		\midrule
		\rowcolor{green!30} \textbf{Ours}     & R+P                                &                                       & \textbf{94.3}                      & \textbf{97.9}                       & \textbf{90.5} & \textbf{91.4} \\
		\bottomrule
	\end{tabular}
	\label{tab1:ntu60}
\end{table*}

\noindent
\textbf{Temporal stride attention.} Several limitations of an attention module exist when the transformer handles the input tokens. In general, the attention module covers all input tokens with scaled-dot products. Thus, the complexity of the attention module is highly dependent on the number of input tokens. In the case of sequential data, this problem is more serious because the input tokens grow with the sizes of the temporal dimensions. Ahn \etal \cite{star} divided temporal dimensions into two groups that contain regularly interleaved tokens. Despite the halved temporal dimensions, the complexity was only slightly reduced, and the temporal correlations of the neighborhood were decoupled. Unlike Ahn \etal \cite{star}, we propose a temporal stride with mitigated complexity and an enhanced temporal correlation for cross-attention. When building input \textit{query}, \textit{key}, and \textit{value} tokens, the temporal dimensions are split into regularly increasing strides to couple various sequential relationships with reduced complexity. As shown in Fig. \ref{fig4} (b), we first set a local time window for a given stride. This window traverses all tokens and specifies the \textit{query}, \textit{key}, and \textit{value} tokens. It not only reduces the number of input tokens of the attention module but also supports temporal representation learning without using all tokens at once.

All of the deformable transformers stated above are repeated $L$-times, as shown in Fig. \ref{fig2} (a). To make the final logits, we concatenate the cross-modal tokens only along with the channel dimension and then feed them to the classification head.

\section{Experiments}
\label{sec:exp}

\noindent
\textbf{Datasets.} We conducted experiments using several representative benchmark datasets: FineGYM \cite{gym}, NTU60 \cite{ntu60}, NTU120 \cite{ntu120}, and PennAction \cite{penn}. FineGYM contains 29K videos with 99 fine-grained action labels collected from gymnastic video clips. NTU60 and NTU120 are representative multimodal datasets that are used for human action recognition. NTU60 consists of 57K videos of 60 action labels collected in a controlled environment. NTU120, which is a superset of NTU60, contains 114K videos of 120 action labels. The NTU datasets use three types of validation protocols following the action subjects and camera settings, \ie, cross-subject (XSub), cross-view (XView), and cross-setup (XSet). In addition, we validated our proposed model with a smaller dataset, PennAction, which contains 2K videos for 15 action labels.

\noindent
\textbf{Settings.} We adopted an AdamW optimizer for 90 epochs with a cosine scheduler with a 5-epoch warm-up. A batch consisted of randomly cropped videos with a pixel resolution of $224\times224$ for training and center-cropped videos for testing. Training and testing were conducted using four NVIDIA Tesla V100 32GB GPUs with APEX.

\begin{table}
	\centering
	\caption{Accuracy comparisons with other SoTA approaches on FineGYM.}
	\begin{tabular}{llcc}
		\toprule
		\multirow{2}{*}{\textbf{Method}}  & \multirow{2}{*}{\textbf{Modality}} & \textbf{Pre-}    & \textbf{Mean} \\
		                                  &                                    & \textbf{trained} & \textbf{(\%)} \\
		\midrule
		ST-GCN \cite{stgcn}               & P$^\dag$                           &                  & 25.2          \\
		TQN \cite{tqn}                    & R                                  & \ding{51}        & 90.6          \\
		PoseC3D \cite{posec3d}            & R+P$^\dag$                         & \ding{51}        & 95.6          \\
		\midrule
		\rowcolor{green!30} \textbf{Ours} & R+P$^\dag$                         &                  & \textbf{90.3} \\
		\bottomrule
	\end{tabular}
	\label{tab2:gym}
\end{table}

\begin{table}
	\centering
	\caption{Accuracy comparisons with other SoTA approaches on PennAction.}
	\begin{tabular}{llcc}
		\toprule
		\multirow{2}{*}{\textbf{Method}}  & \multirow{2}{*}{\textbf{Modality}} & \textbf{Pre-}    & \textbf{Top-1} \\
		                                  &                                    & \textbf{trained} & \textbf{(\%)}  \\
		\midrule
		Pr-VIPE \cite{prvipe}             & P                                  & \ding{51}        & 97.5           \\
		UNIK \cite{unik}                  & P                                  & \ding{51}        & 97.9           \\
		\midrule
		HDM-BG \cite{hdmbg}               & P                                  &                  & 93.4           \\
		3D Deep \cite{3ddeep}             & R+P                                &                  & 98.1           \\
		PoseMap \cite{posemap}            & R+P                                &                  & 98.2           \\
		Multitask CNN \cite{multitaskcnn} & R+P                                &                  & 98.6           \\
		STAR \cite{star}                  & R+P                                &                  & 98.7           \\
		\midrule
		\rowcolor{green!30} \textbf{Ours} & R+P                                &                  & \textbf{	99.7} \\
		\bottomrule
	\end{tabular}
	\label{tab3:penn}
\end{table}

\subsection{Comparison with state-of-the-art approaches}

\noindent
\textbf{NTU60 \& 120.} In action recognition, pre-training for feature extraction and pose information for recognition have a great impact on performance. To obtain objective results in action recognition, it is desirable to use pose information given in the dataset without pre-training. As shown in Table \ref{tab1:ntu60}, PoseC3D \cite{posec3d} shows the best performance at NTU60 and 120, but unlike other methods, it used an estimated pose optimized for recognition models. Therefore, in the experiment, we exclude PoseC3D from quantitative comparison and analysis with SoTA methods.

In Table \ref{tab1:ntu60}, we first compare the top-1 accuracy with various SoTA action recognition methods for the NTU60 dataset. When compared with the TSMF and STAR models trained under fair conditions, our model shows a 0.5\%-2.5\% higher performance, respectively. Among the GNN-based methods, InfoGCN ranks remarkably even when it is trained with a single modality; nonetheless, our model shows higher performance of 94.3\% and 97.9\%, respectively.

We also provide benchmark results against the NTU120 dataset, which has double the number of videos and action labels compared to NTU60. With this result in Table \ref{tab1:ntu60}, the GNN-based SoTA method, InfoGCN, achieved accuracy of 89.8\% and 91.2\% for both protocols. In the case of the multimodal trained method, PoseC3D, it achieved accuracy of 95.3\% and 96.4\% with pre-training and estimated poses. Unlike PoseC3D, the scratch-trained multimodal method, STAR, showed accuracy of 90.3\% and 92.7\%. Finally, our proposed method showed a promising performance under similar conditions, \ie, multimodal and scratch training, with accuracy of 90.5\% and 91.4\%, respectively.

\noindent
\textbf{FineGYM.} In the case of the FineGYM dataset, which is harsher than other datasets because clips have dynamic motions and camera movements from sports games. This dataset has rarely been used for multimodal action recognition, because it does not contain ground-truth skeletons. We used the estimated skeletons from HRNet \cite{hrnet} to apply it to cross-modal learning. Table \ref{tab2:gym} shows the results of the comparison experiments with other SoTA methods for the FineGYM. In the case of a single modality, ST-GCN showed relatively low accuracy because it used the estimated skeletons as our approach. The TQN achieved 90.6\% with the \textit{RGB} single modality. Cross-modality-based methods, including PoseC3D and our approach, showed a relatively high performance compared to other methods. Pre-trained PoseC3D demonstrated a SoTA performance of 95.6\%; however, our method showed promising accuracy despite not applying a pre-training step.

\noindent
\textbf{PennAction.} We validated our method by applying smaller datasets with clearly recorded videos as shown in Table \ref{tab3:penn}. In the GNN regime using the single-pose modality, Pr-VIPE and UNIK achieved high performance of 97.5\% and 97.9\%, respectively. Among cross-modality-based methods, Multitask CNN and STAR show the best performances among the CNN and transformer approaches at 98.6\% and 98.7\%, respectively. Our proposed method outperforms above SoTAs by a large margin of 1.0\%-6.3\%. This result indicates that the proposed method can achieve a good performance regardless of whether small or large datasets are applied.

\begin{figure*}
	\centering
	\includegraphics[width=0.9\linewidth]{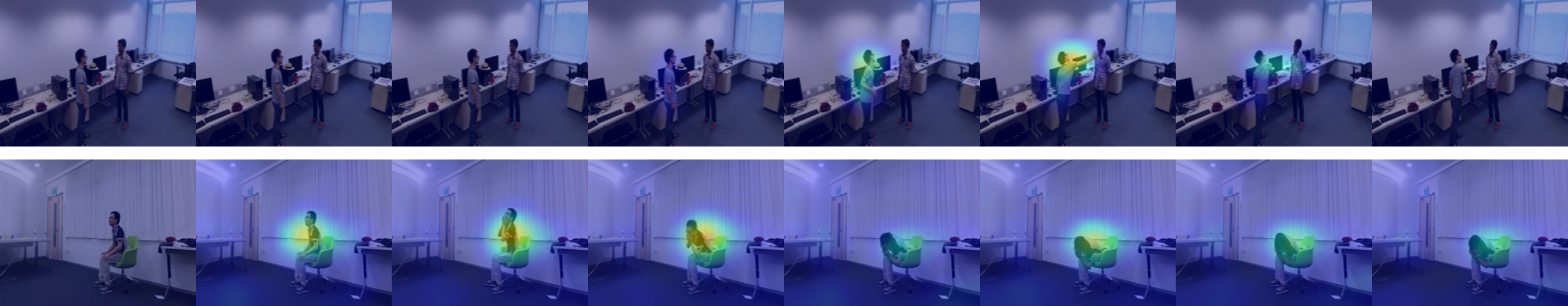}
	\caption{\textbf{Visualized 3D deformable attentions.} The proposed 3D deformable attention found discriminative tokens with strong attention among entire frame sequences. In particular, strong attentions are discovered at only noticeable changes in action.}
	\label{fig7}
\end{figure*}

\begin{figure*}
	\centering
	\includegraphics[width=0.75\linewidth]{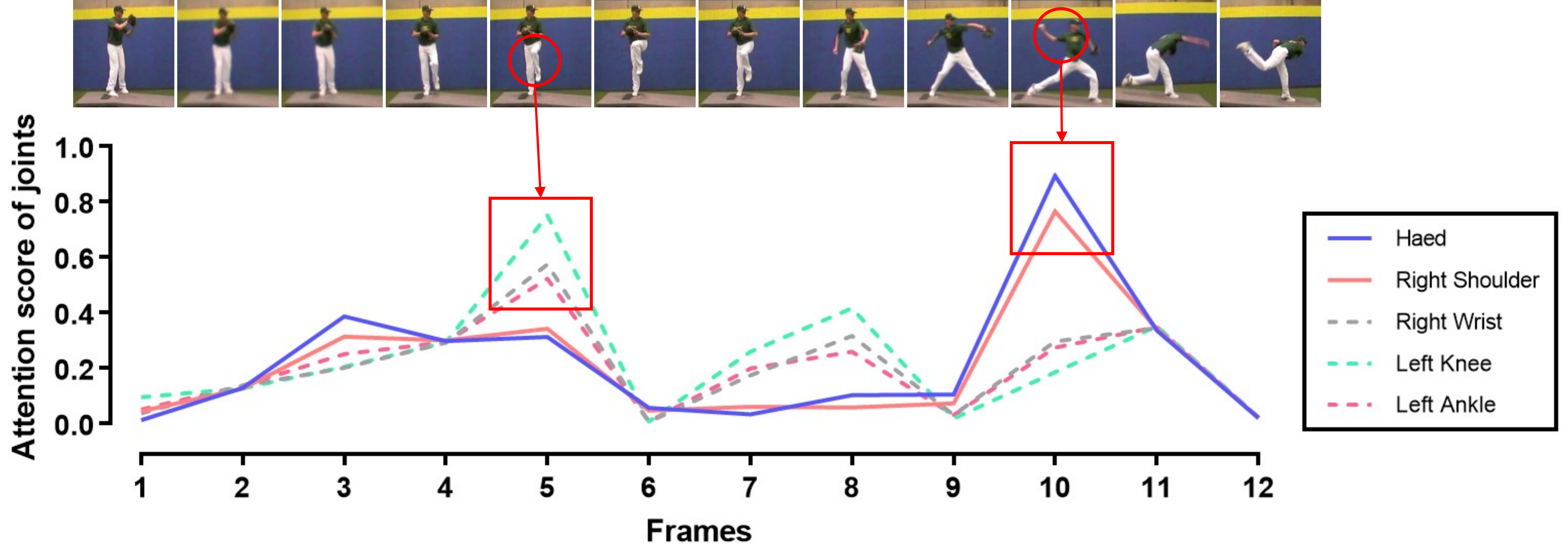}
	\caption{\textbf{Visualized joint stride attentions.} The proposed 3D deformable attention activates the attention score of each joint differently depending on the size of the `\textit{pitching}' action for each frame (shown in color).}
	\label{fig8}
\end{figure*}

\subsection{Qualitative analysis}

\noindent
\textbf{Visualization of 3D deformable attention.} We propose a 3D deformable attention mechanism supporting an adaptive receptive field for action recognition. To demonstrate its effectiveness, we provide qualitative evaluations that show the visualized attention values of selected tokens against different video sequences. As shown in Fig. \ref{fig7}, the proposed method finds intensive regions through a 3D token search of the entire sequence. The activation is relatively low in inactive scenes, but, on the contrary, high activation occurs at a large transition among sequences. In particular, activations appear finely in joints used for actual action and do not appear coarsely. We can confirm that the proposed method not only properly finds tokens that are practically required for recognition in the entire sequence but also shows that strong activation appears in those tokens.

\noindent
\textbf{Visualization of joint stride attention.} To make the proposed model independent of the number of input joints, we propose a local joint stride attention by grouping joint tokens with a sliding window. This approach provides improved efficiency in attention, but we need to verify whether the configured tokens with overlapping contribute to the attention module without full attention. In this experiment, we charted the attention levels of each joint according to sequence frames, as shown in Fig. \ref{fig8}. In practical terms, the joints that move the most when the ball is pitched in the sample video are the head, right hand, and right leg. The chart shows that the attention was largely activated according to this action flow. From this experiment, it is clear that the proposed local joint attention maintains the correlations between the entire joint token and the efficiency.

\begin{table}
	\caption{Ablation study with attention blocks on PennAction.}
	\centering
	\begin{tabular}{ccccc}
		\toprule
		                                 & \textbf{3D def.} & \textbf{Joint} & \textbf{Temp.} & \textbf{Top-1 (\%)} \\
		\midrule
		\ding{172}                       &                  & \ding{51}      & \ding{51}      & 94.8                \\
		\ding{173}                       & \ding{51}        &                &                & 97.9                \\
		\ding{174}                       & \ding{51}        &                & \ding{51}      & 98.2                \\
		\ding{175}                       & \ding{51}        & \ding{51}      &                & 98.5                \\
		\midrule
		\rowcolor{green!30}\textbf{Ours} & \ding{51}        & \ding{51}      & \ding{51}      & \textbf{99.7}       \\
		\bottomrule
	\end{tabular}
	\label{tab4}
\end{table}

\begin{table}
	\caption{Comparison of the Top-1 accuracies for different modal token configurations on PennAction.}
	\centering
	\begin{tabular}{lc}
		\toprule
		\textbf{Modal Tokens}                           & \textbf{Top-1 (\%)} \\
		\midrule
		No tokens                                       & 92.2                \\
		Single token                                    & 97.0                \\
		\rowcolor{green!30} \textbf{Cross-modal tokens} & \bfseries99.7       \\
		\bottomrule
	\end{tabular}
	\label{tab5}
\end{table}

\subsection{Ablation studies}

\noindent
\textbf{Attention modules.} We provide ablation studies on three types of attention modules with PennAction dataset. According to the Table \ref{tab4}, when \ding{172} 3D deformable attention is eliminated from the model, it is confirmed that the accuracy is abruptly degraded ($4.9\%$). Conversely, \ding{173}-\ding{175} other attention modules are found to have relatively less impact. The proposed 3D deformable attention is pivotal for achieving high accuracy in action recognition tasks.

\noindent
\textbf{Cross-modal Tokens.} To analyze the impact of the cross-modal tokens, we conducted ablation studies as shown in Table \ref{tab5}. In the ‘No tokens’ scenario, averaging the last RGB and \textit{pose} tokens led to a performance drop to 92.2\%. Using a single token for all modalities resulted in a slight decrease to 97.0\%. However, our approach achieved the best performance, highlighting the effectiveness of learning multiple modalities in a single transformer design.

\begin{figure}
	\centering
	\begin{subfigure}{0.41\linewidth}
		\includegraphics[width=1.\linewidth]{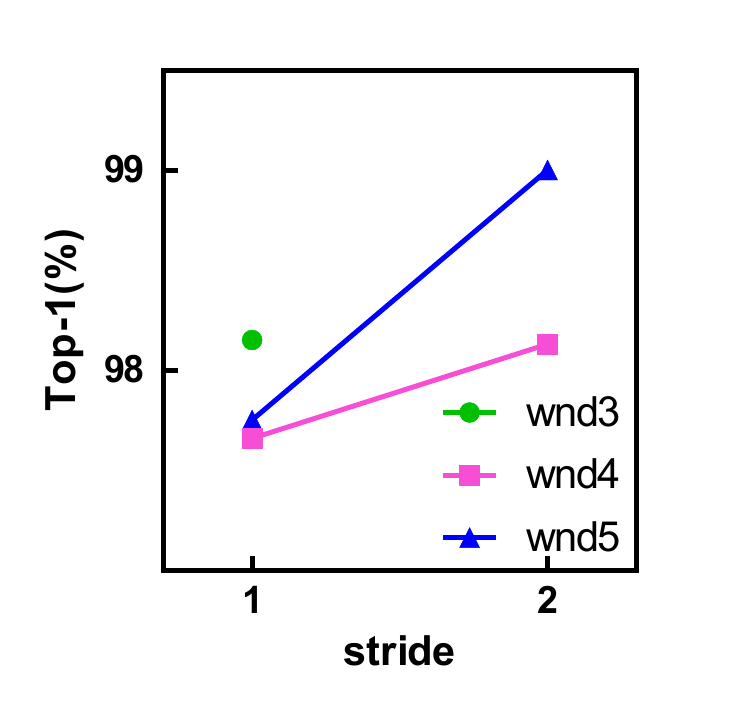}
		\vskip -0.1in
		\caption{}
		\label{fig5:a}
	\end{subfigure}
	\begin{subfigure}{0.55\linewidth}
		\includegraphics[width=1.\linewidth]{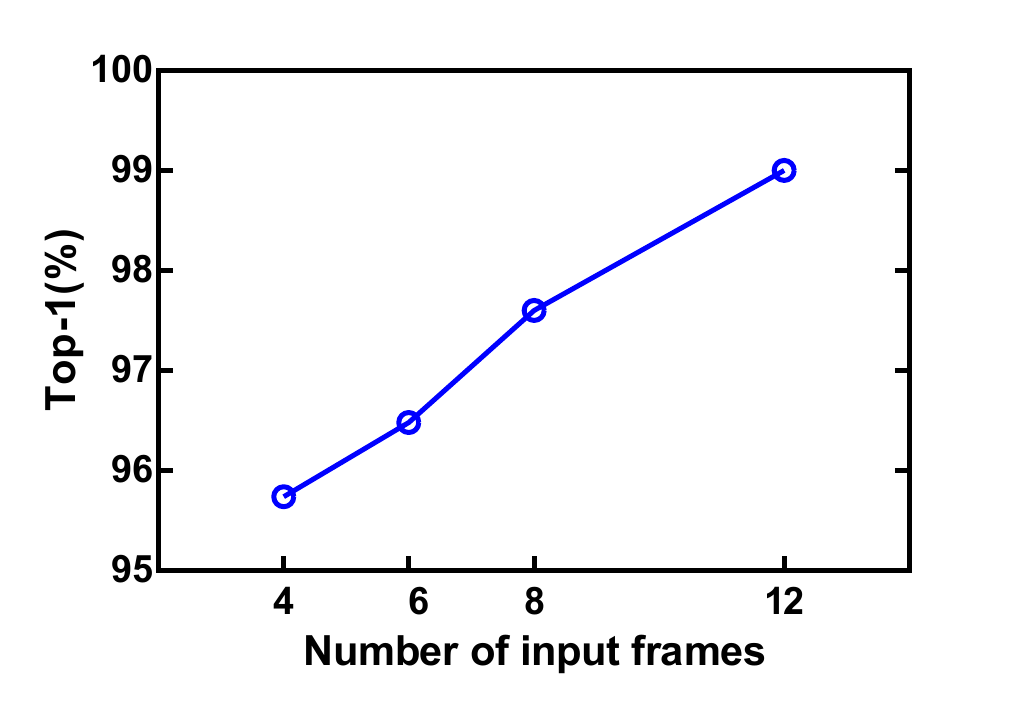}
		\vskip -0.1in
		\caption{}
		\label{fig5:b}
	\end{subfigure}
	\vskip -0.1in
	\caption{(a) Ablation study with different stride on various window ($wnd$) sizes on PennAction. (b) Ablation study with different numbers of input frames during the training phase on PennAction.}
	\label{fig5}
\end{figure}

\noindent
\textbf{Temporal stride.} To solve the limitations of complexity dependent on the number of tokens in an attention module, we propose a local window cross attention using a temporal stride. In this experiment, to evaluate the validity of the proposed approach, we observe the changes in performance by applying various strides to a fixed-sized window using the PennAction dataset. As shown in Fig. \ref{fig5} (a), the best performance is obtained when the stride is about half the size of the window, regardless of the window size. This is because more overlap between temporal tokens fuels the temporal correlations with enhanced efficiency. Therefore, the proposed method faithfully maintains the correlations, even when tokens are divided into local windows.

\noindent
\textbf{Number of input frames.} One of the important factors to determine in the learning of sequential data is the number of input frames. If the number of frames is sufficiently large, better feature representation learning is possible, although the computational cost significantly increases. We therefore empirically determined the optimal number of input frames using the PennAction dataset, as illustrated in Fig. \ref{fig5} (b). In our model, the best performance was achieved using 12 input frames. When the number of frames is smaller than 12, the performance drops slightly, and there is a limit to learning the continuity of the actions.

\section{Conclusion}
\label{sec:conclusion}

ViTs have become the mainstream in various vision tasks, achieving an overwhelming performance; however, it has been used relatively little in action recognition tasks. Therefore, we first proposed a 3D deformable attention consisting of stride window cross attention for better spatiotemporal feature learning, as well as a cross-modal framework for action recognition. The proposed method achieved a newly demonstrated SoTA performance on representative action recognition datasets. Based on the results of qualitative experiments, we can confirm that our proposed method has a strong spatiotemporal feature learning capability for action recognition.

\section*{Acknowledgments}
\thanks{This research was supported by Basic Science Research Program through the National Research Foundation of Korea (NRF) funded by the Ministry of Education(2022R1I1A3058128).}

{\small
	\bibliographystyle{ieee_fullname}
	\bibliography{egbib}
}

\clearpage
\renewcommand{\thetable}{A\arabic{table}}
\renewcommand{\thefigure}{A\arabic{figure}}
\renewcommand{\theequation}{A\arabic{equation}}
\renewcommand{\thealgorithm}{A\arabic{algorithm}}
\setcounter{table}{0}
\setcounter{figure}{0}
\setcounter{equation}{0}
\setcounter{algorithm}{0}
\appendix

\section{Detailed Description of Stride Attentions}

In general, most of the time complexity of transformers is highly related to the attention operation. We offer stride attentions for efficient correlation learning between discretized tokens. In this section, we describe details of the proposed joint stride attention and temporal stride attention.

\subsection{Joint Stride Attention}
The number of \textit{pose} tokens ($\mathrm{P}$) can grow dynamically based on the number of joints ($R$) and people appeared in a scene. Since this is directly related to the amount of computation required for attention, we propose joint stride attention dividing \textit{pose} tokens into several sliding windows. Algorithm \ref{algo1} describes detailed operation of the proposed joint stride attention. According to the notations used in the Algorithm \ref{algo1}, we can organize comparisons of time complexity between full attention and joint stride attention as described in Table \ref{tab1}.

\begin{table}[h]
	\centering
	\caption{Complexity comparisons between full attention and joint stride attention}\label{tab1}
	\begin{tabular}{lc}
		\toprule
		\textbf{Method}        & \textbf{Complexity} \\
		\midrule
		Full attention         & $O(T^2R^2)$         \\
		Joint stride attention & $O(T^2wnd^2)$       \\
		\bottomrule
	\end{tabular}

\end{table}

In joint stride attention, we decompose \textit{pose} tokens using sliding window ($wnd$) with a $stride$ having halved size of $wnd$. If \textit{pose} tokens $\mathrm{P} \in \mathbb{R}^{4C \times T \times R}$ are fed to full attention, the time complexity per layer becomes $O(T^2R^2)$ and the equation is as follows:

\begin{equation}
	\mathbf{O} = [\mathrm{P}||\mathrm{M}_{pose}||\mathrm{M}_{CLS}^{'}]
\end{equation}

\begin{equation}
	\begin{split}
		\mathrm{Full}&\mathrm{Attention}(\mathbf{O}\mathbf{W}_q, \mathbf{O}\mathbf{W}_k, \mathbf{O}\mathbf{W}_v)\\
		& = \sum_{t}^{T}\sum_{r}^{R} \mathrm{softmax}(\frac{\mathbf{O}\mathbf{W}_q^{t,r}\mathbf{O}\mathbf{W}_k^{t,r}}{\sqrt{d_h}}) \mathbf{O}\mathbf{W}_v^{t,r}
	\end{split}
\end{equation}

However, when \textit{pose} tokens are fed to joint stride attention, the time complexity per layer becomes $O(T^2wnd^2)$ and the equation is as follows:

\begin{equation}
	\begin{split}
		\mathrm{Joint}&\mathrm{StrideAttention}(\mathbf{O}\mathbf{W}_q, \tilde{\mathbf{O}}\mathbf{W}_k, \tilde{\mathbf{O}}\mathbf{W}_v)\\
		& = \sum_{t}^{T}\sum_{w}^{wnd}\mathrm{softmax}(\frac{\mathbf{O}\mathbf{W}_q^{t,w}\tilde{\mathbf{O}}\mathbf{W}_k^{t,w}}{\sqrt{d_h}}) \tilde{\mathbf{O}}\mathbf{W}_v^{t,w}
	\end{split}
	\label{eq2}
\end{equation}

\noindent
where $\mathbf{O}$ and $\tilde{\mathbf{O}}$ from Algorithm \ref{algo1}. In Eq. \ref{eq2}, since $wnd$ is always less than $R$, the overall time complexity of joint stride attention is smaller than full attention.

\begin{algorithm}[h]
	\caption{Joint stride attention}\label{algo1}
	\begin{algorithmic}
		\State {\bfseries Input:} \textit{Pose} tokens $\mathrm{P}$, Memorized \textit{CLS} modal token $\mathrm{M}^{'}_{CLS}$, \textit{Pose} modal token $\mathrm{M}_{pose}$, window size $wnd$, \textit{query} weight $\mathbf{W}_q$, \textit{key} weight $\mathbf{W}_k$, \textit{value} weight $\mathbf{W}_v$

		\State {}

		\State $stride \gets \lfloor wnd/2 \rfloor$
		\State $\hat{\mathbf{P}}_{q} \gets \varnothing$ \Comment{\textit{query} set}
		\State $\hat{\mathbf{P}}_{kv} \gets \varnothing$ \Comment{\textit{key} and \textit{value} set}

		\State {}

		\State $i \gets 0$
		\While{$i < (R-wnd)$} \Comment{Split $\mathrm{P}$ into query set $\hat{\mathbf{P}}_{q}$}
		\State $\hat{\mathrm{P}} \gets \mathrm{P}[:, :, i:i+wnd]$ \Comment{$\mathrm{P} \in \mathbb{R}^{4C \times T \times R}$}
		\State $\hat{\mathbf{P}}_{q} \gets \hat{\mathbf{P}}_{q} \cup \hat{\mathrm{P}}$ \Comment{$\hat{\mathrm{P}} \in \mathbb{R}^{4C \times T \times wnd}$}
		\State $i \gets i+stride$
		\If {$i > (R-wnd)$}
		\State $i \gets i-wnd$ \Comment{To assure \textit{query} covers entire tokens}
		\EndIf
		\EndWhile

		\State {}

		\State $i \gets stride$
		\While{$i < (R-wnd)$} \\\Comment{Split $\mathrm{P}$ into \textit{key} and \textit{value} set $\hat{\mathbf{P}}_{kv}$}
		\State $\hat{\mathrm{P}} \gets \mathrm{P}[:, :, i:i+wnd]$ \Comment{$\mathrm{P} \in \mathbb{R}^{4C \times T \times R}$}
		\State $\hat{\mathbf{P}}_{kv} \gets \hat{\mathbf{P}}_{kv} \cup \hat{\mathrm{P}}$ \Comment{$\hat{\mathrm{P}} \in \mathbb{R}^{4C \times T \times wnd}$}
		\State $i \gets i+stride$
		\EndWhile

		\State {}

		\State $D_q \gets |\hat{\mathbf{P}}_q|$ \Comment{$\hat{\mathbf{P}}_q \in \mathbb{R}^{D_q \times 4C \times T \times wnd}$}
		\State $D_{kv} \gets |\hat{\mathbf{P}}_{kv}|$ \Comment{$\hat{\mathbf{P}}_{kv} \in \mathbb{R}^{D_{kv} \times 4C \times T \times wnd}$}

		\State {}

		\Comment{$\mathrm{M}_{pose}, \mathrm{M}^{'}_{CLS} \in \mathbb{R}^{4C \times T \times 1}$}
		\State $\mathbf{O} \gets [\hat{\mathbf{P}}_q || \mathrm{expand}(\mathrm{M}_{pose}, D_q) || \mathrm{expand}(\mathrm{M}^{'}_{CLS}, D_q)]$
		\State $\tilde{\mathbf{O}} \gets [\hat{\mathbf{P}}_{kv} || \mathrm{expand}(\mathrm{M}_{pose}, D_{kv}) || \mathrm{expand}(\mathrm{M}^{'}_{CLS}, D_{kv})]$

		\State $\mathbf{O} \gets \mathbf{O} + \mathrm{MSA}(\mathbf{OW}_q, \tilde{\mathbf{O}}\mathbf{W}_k, \tilde{\mathbf{O}}\mathbf{W}_v)$
		\State $\mathbf{O} \gets \mathbf{O} + \mathrm{FFN}(\mathrm{LN}(\mathbf{O}))$

	\end{algorithmic}
\end{algorithm}

\subsection{Temporal Stride Attention}

Temporal stride attention is proposed to capture temporal changes between each sequential frame and joint. The overall procedure is described in Algorithm \ref{algo2}. The number of input tokens $D_n$  is sum of the number of \textit{RGB}, \textit{pose} and cross modal tokens.  To decompose these tokens into small temporal windows, we apply sliding windows along with the temporal dimension. The complexity comparison result between full attention and temporal stride attention is described in Table \ref{tab2}.

\begin{table}[h]
	\centering
	\caption{Complexity comparisons between full attention and temporal stride attention}
	\begin{tabular}{lc}
		\toprule
		\textbf{Method}           & \textbf{Complexity} \\
		\midrule
		Full attention            & $O(T^2D_n^2)$       \\
		Temporal stride attention & $O(wnd^2D_n^2)$     \\
		\bottomrule
	\end{tabular}
	\label{tab2}
\end{table}

In the case of full attention, the time complexity of attention against the concatenated tokens $\mathbf{N} \in \mathbb{R}^{4C \times T \times D_n}$ becomes $O(T^2D_n^2)$ and the equation is as follows:

\begin{equation}
	\mathbf{N} = [\mathrm{Z} || \mathrm{P} || \mathrm{M}_{RGB} || \mathrm{M}_{pose} || \mathrm{M}_{CLS}]
\end{equation}

\begin{equation}
	\begin{split}
		\mathrm{Full}&\mathrm{Attention}(\mathbf{N}\mathbf{W}_q, \mathbf{N}\mathbf{W}_k, \mathbf{N}\mathbf{W}_v)\\
		& = \sum_{t}^{T}\sum_{n}^{D_n}\mathrm{softmax}(\frac{\mathbf{N}\mathbf{W}_q^{t,n}\mathbf{N}\mathbf{W}_k^{t,n}}{\sqrt{d_h}})\mathbf{N}\mathbf{W}_v^{t,n}
	\end{split}
\end{equation}

On the contrary, when the concatenated tokens are decomposed into several sliding windows, the time complexity per layer becomes $O(wnd^2D_n^2)$ and the equation is as follows:

\begin{equation}
	\begin{split}
		&\mathrm{Temporal}\mathrm{StrideAttention}(\hat{\mathbf{N}}_{q}\mathbf{W}_q, \hat{\mathbf{N}}_{kv}\mathbf{W}_k, \hat{\mathbf{N}}_{kv}\mathbf{W}_v) \\
		& = \sum_{w}^{wnd}\sum_{n}^{D_n}\mathrm{softmax}(\frac{\hat{\mathbf{N}}_{q}\mathbf{W}_q^{w,n}\hat{\mathbf{N}}_{kv}\mathbf{W}_k^{w,n}}{\sqrt{d_h}})\hat{\mathbf{N}}_{kv}\mathbf{W}_v^{w,n}
	\end{split}
\end{equation}

\noindent
where $\hat{\mathbf{N}}_{q}$ and $\hat{\mathbf{N}}_{kv}$ from Algorithm \ref{algo2}. In this case, because $wnd$ is always less than $T$, the overall time complexity of temporal stride attention is smaller than full attention.

\begin{algorithm}[h]
	\caption{Temporal stride attention}\label{algo2}
	\begin{algorithmic}
		\State {\bfseries Input:} \textit{RGB} tokens $\mathrm{Z}$, \textit{Pose} tokens $\mathrm{P}$, \textit{CLS} modal token $\mathrm{M}_{CLS}$, \textit{RGB} modal token $\mathrm{M}_{RGB}$, \textit{Pose} modal token $\mathrm{M}_{pose}$, window size $wnd$, \textit{query} weight $\mathbf{W}_q$, \textit{key} weight $\mathbf{W}_k$, \textit{value} weight $\mathbf{W}_v$

		\State {}

		\State $stride \gets \lfloor wnd/2 \rfloor$
		\State $\hat{\mathbf{N}}_{q} \gets \varnothing$ \Comment{\textit{query} set}
		\State $\hat{\mathbf{N}}_{kv} \gets \varnothing$ \Comment{\textit{key} and \textit{value} set}

		\State {}

		\State $\mathrm{N} \gets [\mathrm{Z} || \mathrm{P} || \mathrm{M}_{RGB} || \mathrm{M}_{pose} || \mathrm{M}_{CLS}]$
		\State $D_n \gets (\frac{H}{8} \times \frac{W}{8} + R + 3)$ \Comment{The number of tokens}

		\State {}

		\State $i \gets 0$
		\While{$i < (T-wnd)$} \Comment{Split $\mathrm{N}$ into \textit{query} set $\hat{\mathbf{N}}_{q}$}
		\State $\hat{\mathrm{N}} \gets \mathrm{N}[:, i:i+wnd, :]$ \Comment{$\mathrm{N} \in \mathbb{R}^{4C \times T \times D_n}$}
		\State $\hat{\mathbf{N}}_{q} \gets \hat{\mathbf{N}}_{q} \cup \hat{\mathrm{N}}$ \Comment{$\hat{\mathrm{N}} \in \mathbb{R}^{4C \times wnd \times D_n}$}
		\State $i \gets i+stride$
		\If {$i > (T-wnd)$}
		\State $i \gets i-wnd$ \Comment{To assure \textit{query} covers entire tokens}
		\EndIf
		\EndWhile

		\State {}

		\State $i \gets stride$
		\While{$i < (T-wnd)$} \\\Comment{Split $\mathrm{N}$ into \textit{key} and \textit{value} set $\hat{\mathbf{N}}_{kv}$}
		\State $\hat{\mathrm{N}} \gets \mathrm{N}[:, i:i+wnd, :]$ \Comment{$\mathrm{N} \in \mathbb{R}^{4C \times T \times D_n}$}
		\State $\hat{\mathbf{N}}_{kv} \gets \hat{\mathbf{N}}_{kv} \cup \hat{\mathrm{N}}$ \Comment{$\hat{\mathrm{N}} \in \mathbb{R}^{4C \times wnd \times D_n}$}
		\State $i \gets i+stride$
		\EndWhile

		\State {}

		\State $\mathbf{N} \gets \hat{\mathbf{N}}_{q} + \mathrm{MSA}(\hat{\mathbf{N}}_{q}\mathbf{W}_q, \hat{\mathbf{N}}_{kv}\mathbf{W}_k, \hat{\mathbf{N}}_{kv}\mathbf{W}_v)$
		\State $\mathbf{N} \gets \mathbf{N} + \mathrm{FFN}(\mathrm{LN}(\mathbf{N}))$

	\end{algorithmic}
\end{algorithm}

\section{Detailed Description of 3D Deformable Attention}\label{app:b}

In this study, we proposed the 3D deformable attention to adaptively capture not only long-term temporal relations but also spatial relations simultaneously. Inspired by Xia \etal \cite{dat}, we rebuilt a deformable attention transformer (DAT) applicable with various video tasks including action recognition. Our proposed method finds discriminative tokens across 3D space while the DAT leverages only 2D space tokens. Details are described in Algorithm \ref{algo3}.

\begin{algorithm}[t]
	\caption{3D deformable attention}\label{algo3}
	\begin{algorithmic}
		\State {\bfseries Input:} \textit{RGB} tokens $\mathrm{Z}$, \textit{CLS} modal token $\mathrm{M}_{CLS}$, \textit{RGB} modal token $\mathrm{M}_{RGB}$, \textit{query} weight $\mathbf{W}_q$, \textit{key} weight $\mathbf{W}_k$, \textit{value} weight $\mathbf{W}_v$, kernel size $k$, 3D conv block $f_{off}$, bilinear sampling function $g$, trainable parameter $\omega$


		\Function{3DTS}{$\mathrm{Z}; \omega$}
		\State $\mathrm{Z} \gets \mathrm{reshape}(\mathrm{Z})$
		\Comment{$\mathrm{Z} \in \mathbb{R}^{4C \times T \times {H\over8} \times {W\over8}}$}
		\State $\Delta p \gets \mathrm{tanh}(f_{off}(\mathrm{Z};\omega))$
		\Comment{$\Delta p \in \mathbb{R}^{3\times\tilde{T}\times\tilde{H}\times\tilde{W}}$}
		\State $p \gets$ reference points from 3D grid
		\\
		\Comment{$p \in \mathbb{R}^{3\times\tilde{T}\times\tilde{H}\times\tilde{W}}$}

		\State Initialize $\tilde{\mathrm{Z}} \in \mathbb{R}^{4C \times T \times {H\over8} \times {W\over8}}$
		\For{$(p_x, p_y, p_z) \in p+\Delta p$}
		\State $\tilde{\mathrm{z}} \gets 0$
		\For{$(r_x, r_y, r_z) \in [\{1...{W\over8}\}, \{1...{H\over8}\}, \{1...T\}]$}\\
		\Comment{get spatiotemporal coordinates $r_{\{x,y,z\}}$}
		\State $\phi \gets g(p_x, r_x)g(p_y, r_y)g(p_z, r_z)$
		\State $\tilde{\mathrm{z}} \gets \tilde{\mathrm{z}} +  \phi \mathrm{Z}[:, r_z, r_y, r_x]$
		\EndFor

		\State $\tilde{\mathrm{Z}}[:, p_z, p_y, p_x] \gets \tilde{\mathrm{z}}$
		\EndFor

		\State $\mathbf{return}$ $\mathrm{flat}(\tilde{\mathrm{Z}})$
		\EndFunction

		\State Initialize 3D conv. parameter ($\omega$) with $k$
		\State $\tilde{\mathrm{Z}} \gets \mathrm{3DTS}(\mathrm{Z}; \omega)$

		\State $\mathbf{X}, \tilde{\mathbf{X}} \gets [\mathrm{Z} || \mathrm{M}_{RGB} || \mathrm{M}_{CLS}], [ \tilde{\mathrm{Z}} || \mathrm{M}_{RGB} || \mathrm{M}_{CLS}]$

		\State $\mathbf{X} \gets \mathbf{X} + \mathrm{MSA}(\mathbf{X}\mathbf{W}_q, \tilde{\mathbf{X}}\mathbf{W}_k, \tilde{\mathbf{X}}\mathbf{W}_v)$
		\State $\mathbf{X} \gets \mathbf{X} + \mathrm{FFN}(\mathrm{LN}(\mathbf{X}))$
	\end{algorithmic}
\end{algorithm}

\begin{figure}
	\centering
	\includegraphics[width=\linewidth]{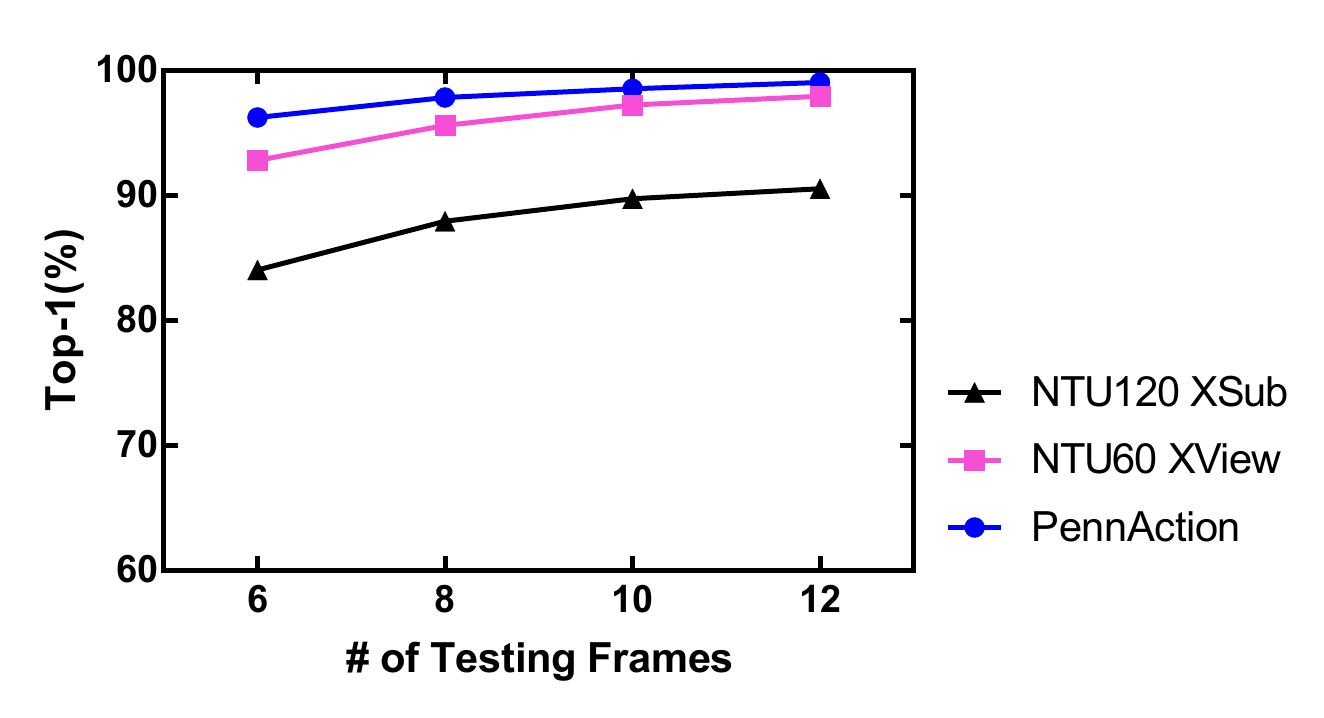}
	\vskip -0.1in
	\caption{Ablation study with different numbers of frames during test phase against model trained with 12 frames.}
	\label{fig6}
	\vskip -0.18in
\end{figure}

\section{On the fly frames in test phase.}
The proposed method showed a good performance on several benchmarks using the suggested modules for capturing temporal changes. According to the Fig. 5 (b) in the paper, it was observed that the performance has a high relevance for the number of input frames in training phase. For that reason, we assumed that if the model well captures spatiotemporal relations on dense frame condition in training, then it will be able to defense degradation of performance on sparse test frames. In practical application, some models may have to be run in sparse frames due to environmental limitations. We provided the evaluation results by diversifying the number of input frames in a model trained using 12 frames. The results verify the power of the spatiotemporal feature learning of the proposed method. In Fig. \ref{fig6}, the proposed method shows a uniform performance for various numbers of input frames. Therefore, the proposed method is robust in learning spatiotemporal features, even if the number of testing frames is sparse.

\section{Additional Qualitative Results}

\subsection{Additional Joint Stride Attention Visualization}
\begin{figure*}[h]
	\centering
	\begin{subfigure}{0.85\linewidth}
		\includegraphics[width=\linewidth]{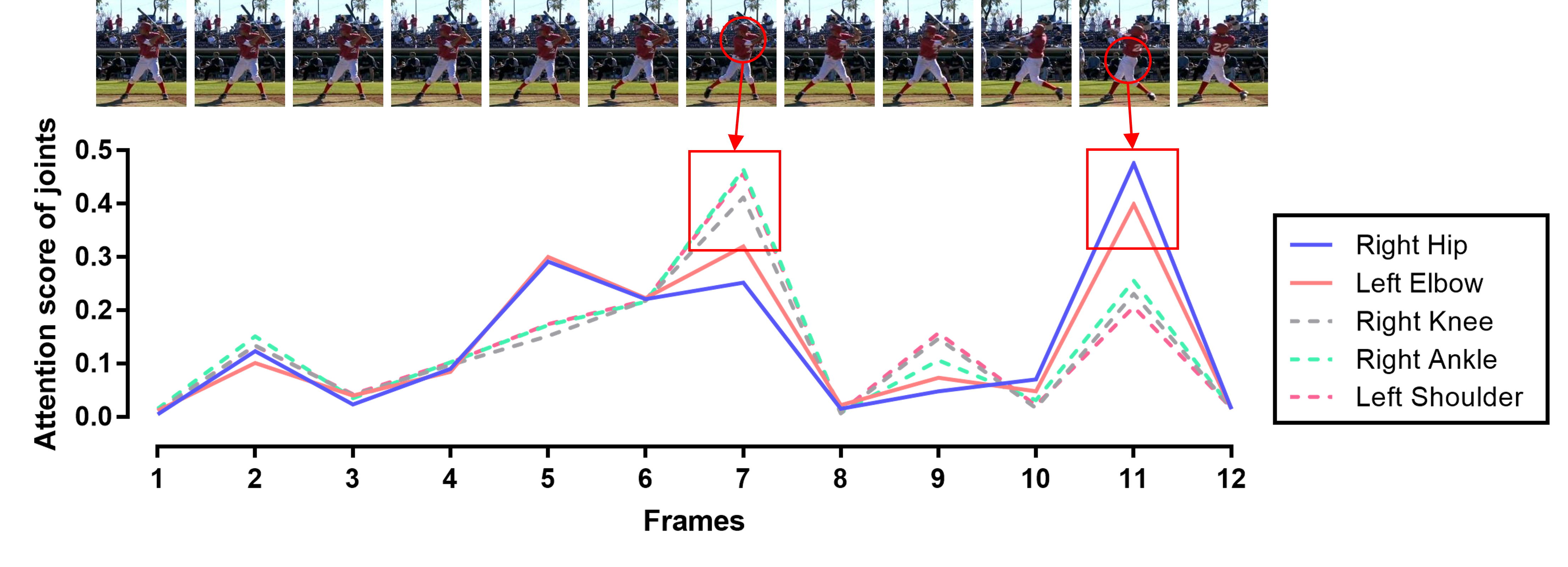}
		\caption{baseball swing}
		\vspace{1em}
	\end{subfigure}

	\begin{subfigure}{0.85\linewidth}
		\includegraphics[width=\linewidth]{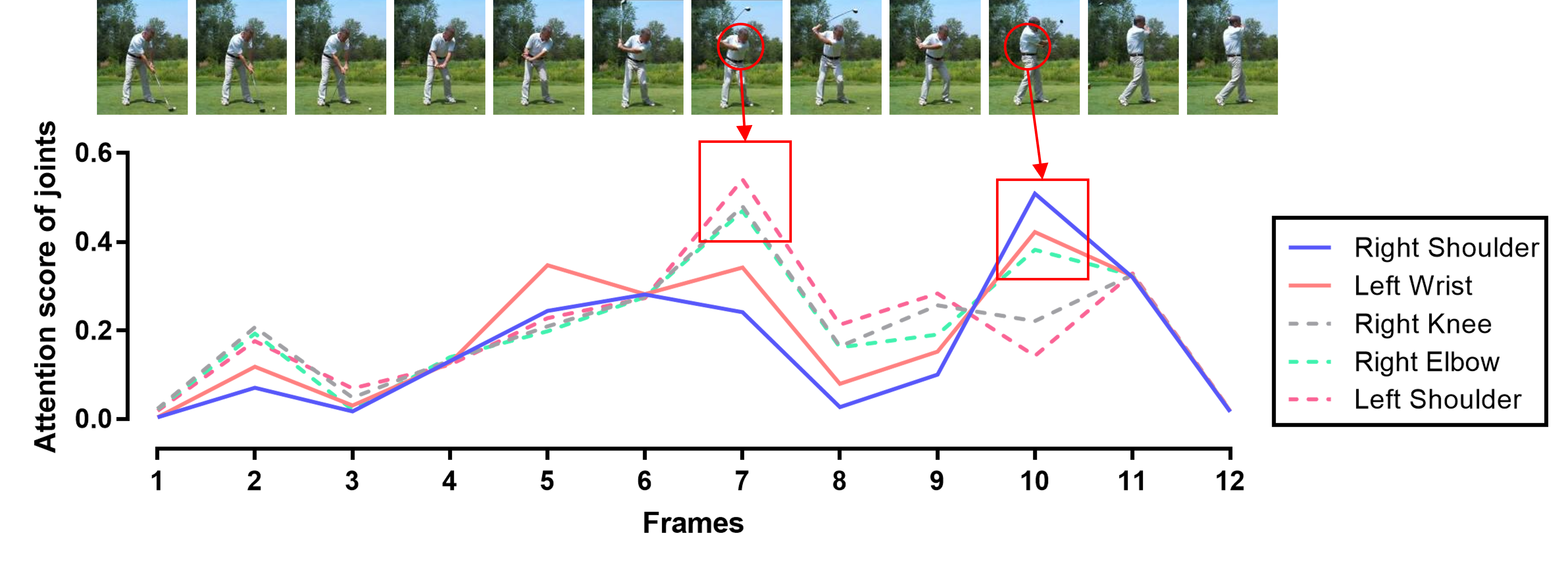}
		\caption{golf swing}
		\vspace{1em}
	\end{subfigure}

	\begin{subfigure}{0.85\linewidth}
		\includegraphics[width=\linewidth]{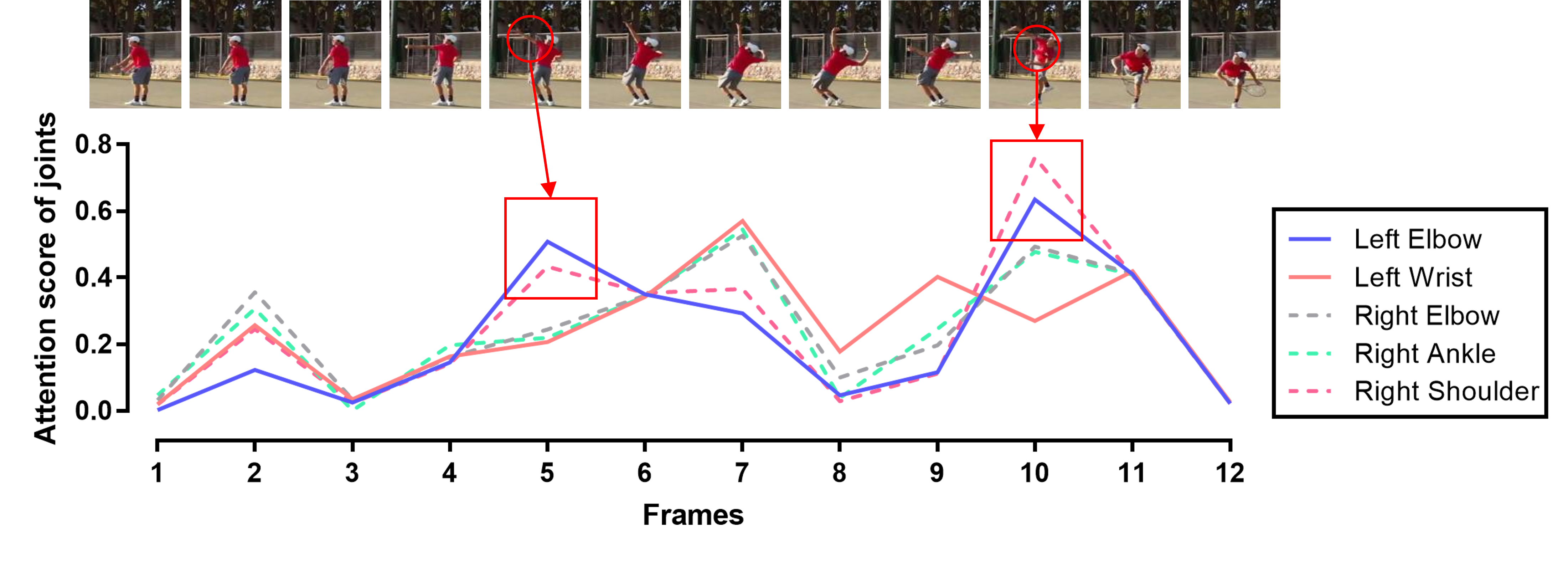}
		\caption{tennis serve}
		\vspace{1em}
	\end{subfigure}
	\caption{Visualization of joint stride attention on PennAction}
	\label{figjoint}
\end{figure*}

We present the result of analyzing the role of each \textit{pose} token in cross-modal learning. The visualizations of each \textit{pose} token for more diverse actions are shown in Fig. \ref{figjoint}.

\subsection{Additional 3D Deformable Attention Visualization}
\begin{sidewaysfigure*}[t!]
	\centering
	\begin{subfigure}{\linewidth}
		\includegraphics[width=\linewidth]{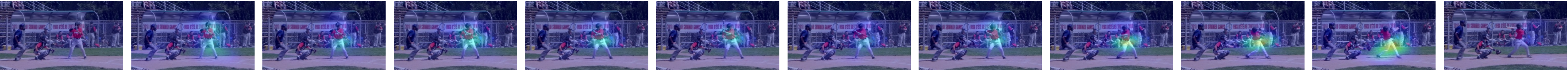}
		\caption{baseball swing}
	\end{subfigure}
	\begin{subfigure}{\linewidth}
		\includegraphics[width=\linewidth]{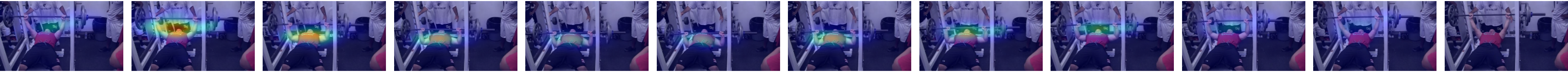}
		\caption{bench press}
	\end{subfigure}
	\begin{subfigure}{\linewidth}
		\includegraphics[width=\linewidth]{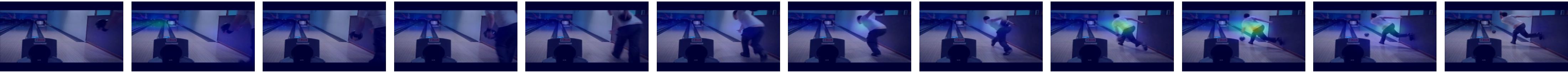}
		\caption{bowling}
	\end{subfigure}
	\begin{subfigure}{\linewidth}
		\includegraphics[width=\linewidth]{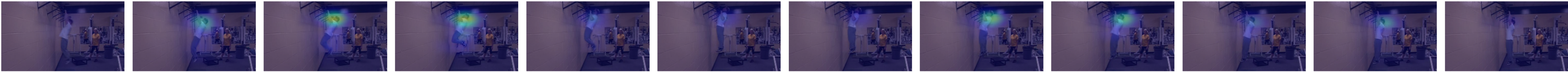}
		\caption{pull ups}
	\end{subfigure}
	\begin{subfigure}{\linewidth}
		\includegraphics[width=\linewidth]{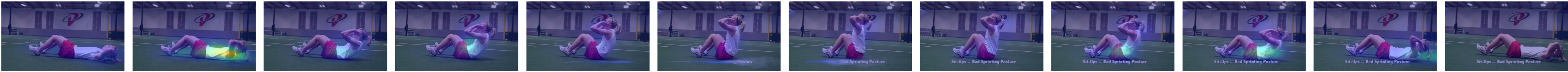}
		\caption{sit ups}
	\end{subfigure}
	\begin{subfigure}{\linewidth}
		\includegraphics[width=\linewidth]{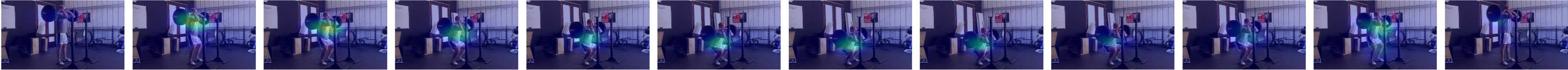}
		\includegraphics[width=\linewidth]{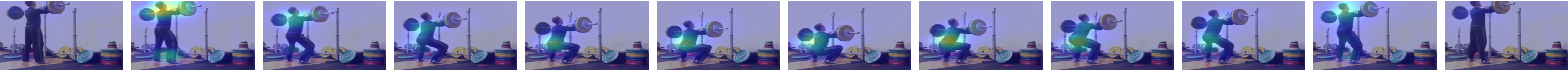}
		\caption{squats}
	\end{subfigure}
	\caption{Visualization of 3D deformable attention on PennAction}
	\label{figpenn}
\end{sidewaysfigure*}

\begin{sidewaysfigure*}[t!]
	\centering
	\begin{subfigure}{\linewidth}
		\includegraphics[width=\linewidth]{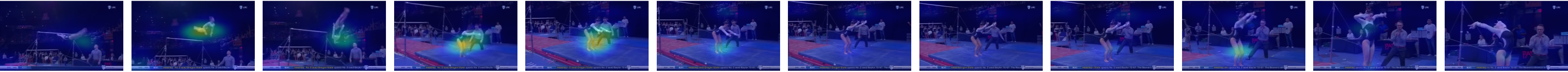}
		\caption{(swing forward) double salto backward stretched}
	\end{subfigure}
	\begin{subfigure}{\linewidth}
		\includegraphics[width=\linewidth]{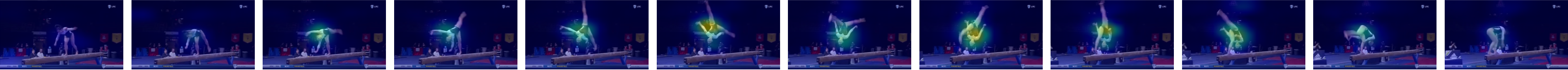}
		\caption{free aerial cartwheel landing in cross position}
	\end{subfigure}
	\begin{subfigure}{\linewidth}
		\includegraphics[width=\linewidth]{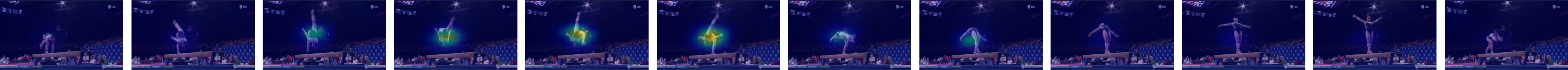}
		\caption{free aerial walkover forward, landing on one or both feet}
	\end{subfigure}
	\begin{subfigure}{\linewidth}
		\includegraphics[width=\linewidth]{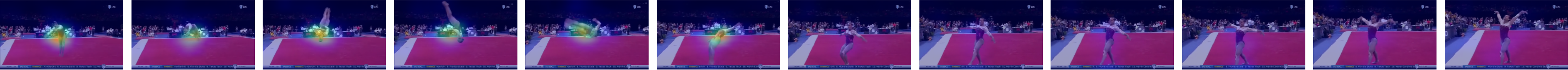}
		\caption{salto forward stretched with 1 twist}
	\end{subfigure}
	\begin{subfigure}{\linewidth}
		\includegraphics[width=\linewidth]{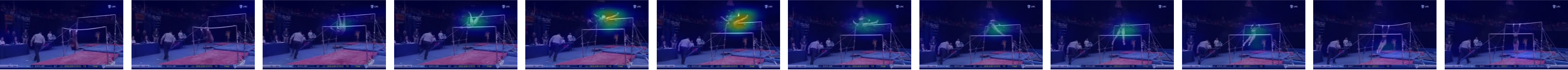}
		\caption{transition flight from low bar to high bar}
	\end{subfigure}
	\begin{subfigure}{\linewidth}
		\includegraphics[width=\linewidth]{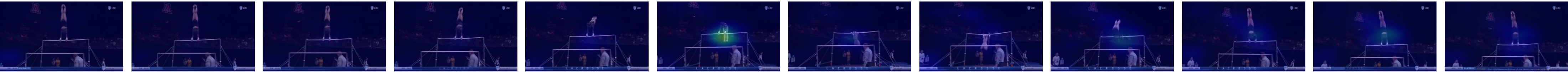}
		\caption{pike sole circle backward to handstand}
	\end{subfigure}
	\begin{subfigure}{\linewidth}
		\includegraphics[width=\linewidth]{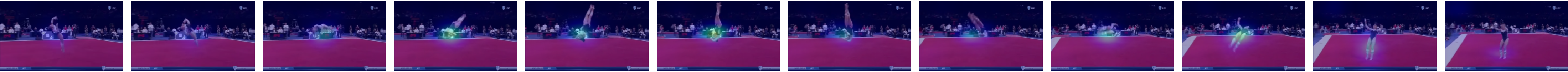}
		\caption{salto backward stretched with 1.5 twist}
	\end{subfigure}
	\begin{subfigure}{\linewidth}
		\includegraphics[width=\linewidth]{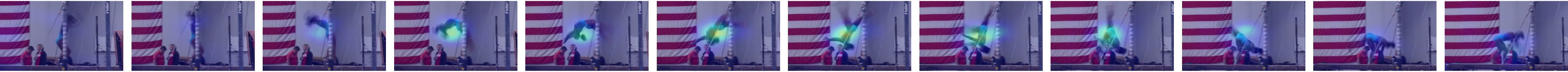}
		\caption{salto backward stretched-step out (feet land successively)}
	\end{subfigure}
	\begin{subfigure}{\linewidth}
		\includegraphics[width=\linewidth]{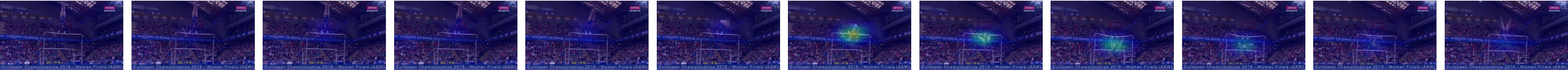}
		\caption{stalder backward to handstand}
	\end{subfigure}
	\caption{Visualization of 3D deformable attention on FineGYM}
	\label{figgym}
\end{sidewaysfigure*}

\begin{sidewaysfigure*}[t!]
	\centering
	\begin{subfigure}{\linewidth}
		\includegraphics[width=\linewidth]{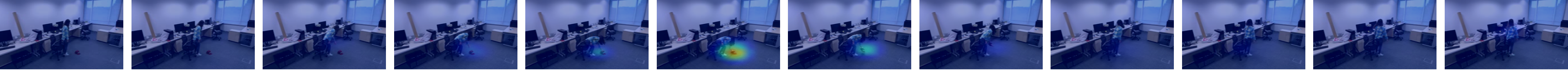}
		\caption{put on a shoe}
	\end{subfigure}
	\begin{subfigure}{\linewidth}
		\includegraphics[width=\linewidth]{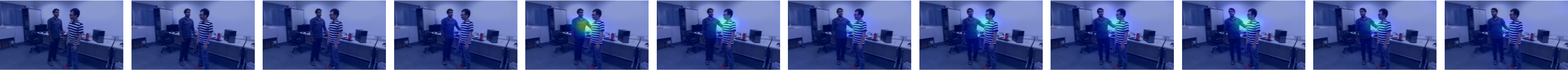}
		\caption{pat on back}
	\end{subfigure}
	\begin{subfigure}{\linewidth}
		\includegraphics[width=\linewidth]{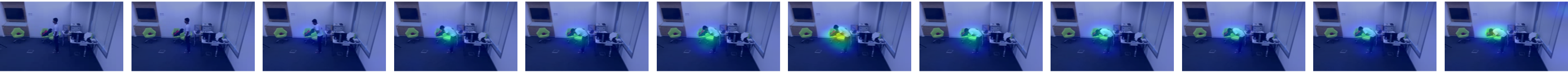}
		\caption{nausea/vomiting}
	\end{subfigure}
	\begin{subfigure}{\linewidth}
		\includegraphics[width=\linewidth]{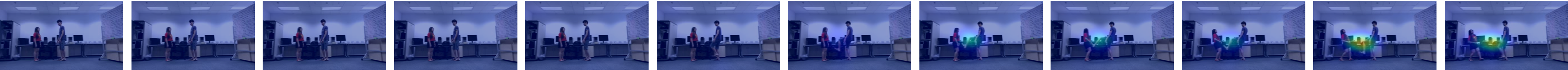}
		\caption{kicking}
	\end{subfigure}
	\begin{subfigure}{\linewidth}
		\includegraphics[width=\linewidth]{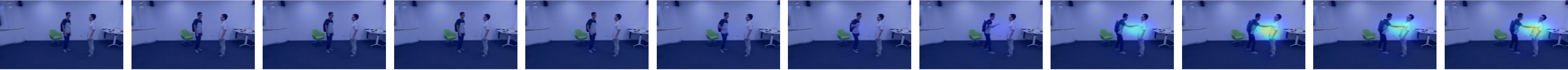}
		\includegraphics[width=\linewidth]{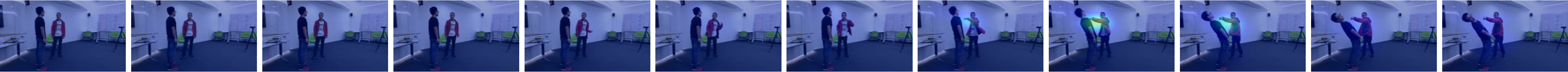}
		\caption{punch/slap}
	\end{subfigure}
	\begin{subfigure}{\linewidth}
		\includegraphics[width=\linewidth]{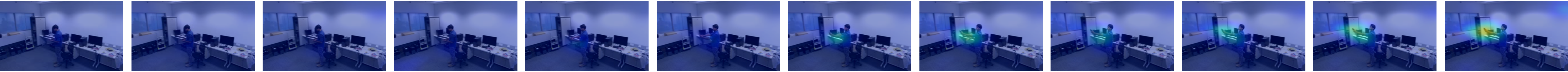}
		\caption{taking a selfie}
	\end{subfigure}
	\begin{subfigure}{\linewidth}
		\includegraphics[width=\linewidth]{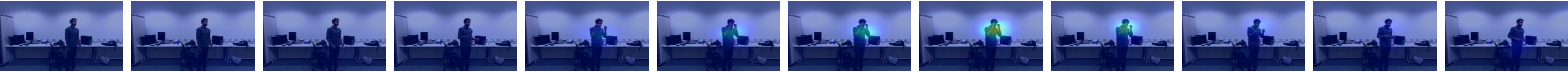}
		\caption{take off glasses}
	\end{subfigure}
	\begin{subfigure}{\linewidth}
		\includegraphics[width=\linewidth]{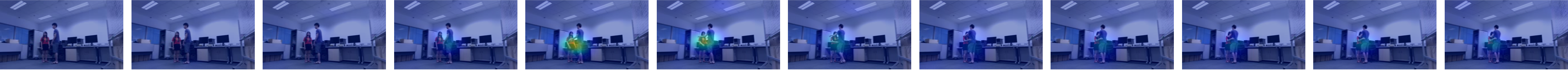}
		\caption{hugging}
	\end{subfigure}
	\begin{subfigure}{\linewidth}
		\includegraphics[width=\linewidth]{suppfigs/ntu_pick_up.png}
		\caption{pick up}
	\end{subfigure}
	\caption{Visualization of 3D deformable attention on NTU120}
	\label{figntu}
\end{sidewaysfigure*}

In this section, we provide additional visualization of 3D deformable attention for more diverse actions on each dataset. In the case of PennAction, attention is accurately appeared to the person who is the subject of the action even in complex backgrounds as shown in Fig. \ref{figpenn}, and it can be seen that attention is occurring intensively in the frame representing the action. In terms of FineGYM, it consists of fine-grained gymnastic frames with dynamic camera moving. Our proposed 3D deformable attention accurately tracks gymnasts performing dynamic movements as shown in Fig. \ref{figgym}, and clearly understands the differences in each fine-grained actions, even for actions using the same equipment but with different labels. In the case of NTU120, which has a relatively simple background, the proposed method accurately finds key elements for various actions. The interaction between two people is also well tracked, especially in the case of the `\textit{pick up}' action label, the actor's action is important, but the fact that there is a dropped object may be more important to accurately classify this action as shown in Fig. \ref{figntu}.

\end{document}